\documentclass[shortAfour,sageh,times]{sagej_no_sage}

\usepackage{array}
\usepackage{amsmath}
\usepackage{amsfonts}
\usepackage{amssymb}
\usepackage{paralist}
\usepackage{graphicx}
\usepackage{multirow}
\usepackage[caption=false,font=footnotesize]{subfig}

\graphicspath{ {figures/} }
\usepackage[ruled]{algorithm2e}
\usepackage{algpseudocode} 
\newcommand{\vect}[1]{(#1_1,#1_2,\dots,#1_n)}
\DeclareMathOperator*{\argmin}{arg\,min}
\DeclareMathOperator*{\argmax}{arg\,max}

\newlength\mylen
\newcommand\myinput[1]{%
  \settowidth\mylen{\KwIn{}}%
  \setlength\hangindent{\mylen}%
  \hspace*{\mylen}#1\\}

\runninghead{Francis et al.}

\title{Occupancy Map Building through Bayesian Exploration}

\author{Gilad Francis, Lionel Ott, Roman Marchant and Fabio Ramos}

\affiliation{School of Information Technologies, University of Sydney, Australia}

\corrauth{Gilad Francis, 
	School of Information Technologies,
	University of Sydney,
	NSW 2006, 
	Australia.}

\email{gfra8070@uni.sydney.edu.au}

\begin{document}

\bibliographystyle{SageH}

\title{Occupancy Map Building through Bayesian Exploration}

\begin{abstract}
We propose a novel holistic approach for safe autonomous exploration and map building based on constrained Bayesian optimisation. 
This method finds optimal continuous paths instead of discrete sensing locations that inherently satisfy motion and safety constraints. 
Evaluating both the objective and constraints functions requires forward simulation of expected observations. As such evaluations are costly, the Bayesian optimiser proposes only paths which are likely to yield optimal results and satisfy the constraints with high confidence. By balancing the reward and risk associated with each path, the optimiser minimises the number of expensive function evaluations. We demonstrate the effectiveness of our approach in a series of experiments both in simulation and with a real ground robot and provide comparisons to other exploration techniques. Evidently, each method has its specific favourable conditions, where it outperforms all other techniques. Yet, by reasoning on the usefulness of the entire path instead of its end point, our method provides a robust and consistent performance through all tests and performs better than or as good as the other leading methods.
\end{abstract}


\maketitle

\section{Introduction}

Autonomous exploration is a challenging dynamic decision-making process, where the goal is to build a representation of an initially unknown environment. While exploring, the robot determines its own position and decides where to move next based on its task. Ideally, these decisions correspond to continuous trajectories which determine the goal pose of the robot and a path leading to that point. Such a path would maximise the robot's objective, while ensuring safety. However, given the dimensionality and shape of the search space imposed by the motion constraints, a closed-form solution to the general exploration problem is intractable. The numerous exploration techniques available in the literature provide different approaches for making a decision on which path to follow, each with its own strategy for dealing with the uncertainty of the model and solution horizon. Risk and safety are typically addressed during execution of the chosen path, and not as an integral part of the exploration decision-making process.

In this paper, we present a novel framework for autonomous exploration over continuous paths using constrained \textit{Bayesian optimisation} (BO) \citep{Gelbart2014}, which we term \textit{constrained Bayesian exploration} (CBE). 
\textit{A-priori}, the exploration objective and constraints functions are unknown, i.e. have no closed-form expression. Knowledge about any of these functions is obtained solely from noisy observations, typically using forward simulation. A naive approach for optimising path decision in such a case is to employ an exhaustive search. Of course, such a process is computationally infeasible due to the size of the search space and cost associated with evaluating the reward and constraints functions over entire paths. BO uses a completely different approach, which turns autonomous exploration into an active learning process over the continuous search space. Instead of exhaustive sampling, BO learns probabilistic surrogate models for the expensive-to-evaluate reward and constraints functions \citep{Brochu2010}. A simple and cheap heuristic function, called acquisition function, guides an efficient sampling schedule based on the posterior mean and variance of the surrogate models. The acquisition function balances the exploration-exploitation trade-off which guarantees convergences to the the optimum while ensuring probabilistic completeness of the objective and constraints functions models.
   
The main contributions of this paper are; First, we formulate the problem of building an occupancy map of an unknown environment 
as an optimisation problem in a constrained continuous action domain. Second, we develop CBE, an innovative method to solve this problem, i.e. optimise path decision whilst keeping the robot safe and within its dynamic constraints.

The remainder of this paper is organised as follows. Section \ref{sec:RelatedWork} surveys the work related to autonomous exploration. Section \ref{sec:explo} provides background on combined exploration and map building. Section \ref{sec:CBO} gives an introduction of the basic building blocks of constrained Bayesian exploration and details the algorithms behind CBE. Experimental results and analysis for various scenarios are shown in Section \ref{sec:experimental}. Finally, Section \ref{sec:conclusions} draws conclusions on the proposed method.

\section{Related Work}
\label{sec:RelatedWork}
Autonomous exploration can be seen as an active learning process aimed at minimising uncertainty and producing 
high-fidelity maps \citep{Makarenko2002,Stachniss2009}. Exploration requires solving simultaneously mapping, path planning and localisation. Due to its complexity, existing research has mainly focused on solving a relaxed form of this problem, by either decoupling processes or by limiting the solution search space.
 
The plethora of autonomous exploration methods is categorically divided into two branches; frontier-driven and information-theoretic. 
A quantitative comparison between the various exploration algorithms is presented in \citet{Julia2012}.

The key concept in any frontier-based exploration is moving towards the edges of the known space, i.e. the boundary between free space 
and unmapped regions \citep{Yamauchi1997}. In its simplest form, after identifying and clustering frontiers, the robot moves towards 
the closest one. Other authors suggest various utility functions to prioritise candidate frontier locations. \citet{Gonzalez-Banos2002} used expected information gain at the frontier and travelling cost. \citet{Basilico2011} incorporated the overlap with known space as a measure for self-localisation. Extensions for 3D autonomous exploration have also been suggested by various authors \citep{Dornhege2013,Shen2012,Shade2011}.

Information driven exploration strategies minimise a utility associated with the uncertainty of the map. Early work dealt with only finding the \textit{next best view} (NBV), which is the discrete location that will have the greatest effect on the utility function. \citet{Whaite1997} 
proposed minimising the entropy of the map. \textit{Mutual information} (MI) has also been suggested as a measure for the predicted 
reduction of map uncertainty \citep{Elfes1996,Bourgault2002}. \citet{Julian2014} suggested MI to encode geometric dependencies and drive exploration 
into unexplored regions in a similar fashion to frontier based approaches. \citet{Makarenko2002} proposed an integrated 
exploration that combines the goal functions of map and localisation uncertainties with the cost of navigation, balancing exploration with \textit{simultaneous localisation and mapping} (SLAM) loop-closures. \citet{Tovar2006} extended this technique by selecting several observation points using a tree search. Yet, decision on the path passing through these points is still not part of the optimisation process. \citet{Stachniss2005} used a particle filter to calculate the expected information gain of an action. However, this formulation still defines point actions of either loop-closure or exploration. A method to generate a path based on information potential fields is proposed by \citet{Vallve2015}. The path is generated by applying a grid-step gradient on the potential fields, hence the resulting path does not necessarily complies with the robot kinematic restriction.  

Several non-myopic exploration methods have emerged in recent years. These methods treat exploration as a sequential decision process. 
\citet{Yang2013} used a \textit{rapidly-exploring random tree} (RRT) planner with Gaussian process occupancy map to generate a safe path that minimises 
MI. Similarly to CBE, the RRT planner does not set a goal point but rather explores promising paths. The difference in algorithms lies in the optimisation process. While RRT uses predefined valid branches for its tree, CBE optimises path selection over the continuous domain. Furthermore, CBE learns the path constraints, which makes it a more flexible algorithm. \citet{Charrow-RSS-15} combined both frontier and information-theoretic approaches. Global goal candidates are produced 
by identifying frontiers. A coarse path to each candidate is generated using local motion primitives that satisfy the kinematic envelope of 
the robot. After assessing the information gain of all candidates, the best path is refined by a local gradient-based optimiser. While this method optimises control inputs in continuous space, the search is limited to a single promising path. CBE, on the other hand, does not define a goal point, and its optimisation is done on the entire domain of controls. \citet{Kollar2008} proposed an exploration procedure that maximises map coverage, by choosing a set of observation points that the robot trajectory 
must pass through. The executed path minimises the errors introduced by the robot motion. The control policy is implemented by a \textit{support vector 
machine} (SVM) classifier trained off-line. Another exploration approach was introduced by \citet{Marchant2014}. In this case, Bayesian optimisation was used to learn and optimise a utility function along continuous paths. They employed two instantiations of BO, one for the probabilistic model of the utility and another to select a continuous informative path. However, their optimisation process does not consider any motion or safety constraints, which are learned and incorporated in the CBE framework. In a more recent work, \citet{Marchanta} developed a sequential Bayesian optimisation method within a \textit{partially observable Markov decision process} (POMDP) framework. They used \textit{Monte Carlo tree search} (MCTS) to approximate the unconstrained solution for a spatio-temporal monitoring problem. 
\citet{Martinez-Cantin2007} and more recently \citet{Martinez-Cantin2009} utilised Bayesian optimisation to find control policies that minimise the state error of a robot and landmarks. While they used a different cost function, their method resembles the approach taken in this work. However, CBE extends this method by incorporating and learning unknown constraints during the optimisation. \citet{Lauri2015a} used a POMDP with a MI objective to plan exploration paths with a fixed horizon using tree search methods to optimise the exploration policy. Their method relies on a \textit{Monte-Carlo} (MC) approximation for the MI objective forward simulation. CBE, on the other hand, uses BO to efficiently manage the objective and constraints functions sampling. Another POMDP continuous-domain planning technique was presented by \citet{Indelman2015}. They treated the exploration as an optimisation of the expected cost over several look-ahead steps. The cost at each step is inferred from the joint probability distribution of the robot and environment state. As the expected cost has a closed form expression, the authors use gradient descent to locally optimise the policy selection. A similar approach was taken by \citet{Rafieisakhaei2016} which defined a penalised cost function based on the \textit{maximum-a-posteriori} (MAP) state estimate and control effort penalties. Our method takes a different approach to optimisation. Instead of using a closed-form expression for the objective, CBE learns it from samples, which results in a more flexible solution. Therefore, there is no need to define an expression for the objective and constraints. Rather the non-parametric structure captures our belief about the cost function. Coupling that with the BO framework, provides better guarantees that a solution will converge to the global optimum.

In summary, in this paper we take a more holistic approach to exploration. Ideally, one would like to select the path that yields the highest reward. However, evaluating path reward is expensive and as such, simple global optimisation strategies are rendered impractical. Therefore, most exploration techniques break this problem into two sub-problems; finding the next observation point(s) and selecting a path through these points. By contrast, our method uses a modified version of BO that finds a solution for these two sub-problems at the same time. The reward function and any associated motion constraints, such as turn rate limits or obstacles, are treated as functions which are learned by the optimiser. The output of the optimisation procedure is a path that maximises the reward without violating any of the constraints. 

\section{Exploration and Map Building}
\label{sec:explo}
In this section we formally describe the problem of safe autonomous exploration for building occupancy maps. 
We start by describing the map representation before formulating the exploration process as an optimisation problem over the robot's action space.
\subsection{Occupancy Grid Maps}
The aim of autonomous exploration is to map an unknown environment. Formally, mapping is an inference process where given 
a set of observations, $\boldsymbol{z}$, taken at known poses, $\boldsymbol{x}$, the posterior distribution over maps, $\boldsymbol{m}$ is given by $p(\boldsymbol{m}|\boldsymbol{z},\boldsymbol{x})$.
Using grid maps, the complexity of calculating the posterior can be reduced ~\citep{Thrun2005a}. In a grid map, the 2D world is discretised 
into cells, where each grid cell, $m_i$, is an independent Bernoulli random variable. This assumption simplifies calculations since the map posterior is now a product of the individual cells:
\begin{equation}\label{fig:general_map}
	p(\boldsymbol{m}) = \prod_{i}p(m_i).
\end{equation}

It is worth noting that while the exploration algorithm drives the process of map building, updating the map with new observations is controlled by a separate and external routine; e.g. gMapping \citep{Grisetti2007}. As such, the use of occupancy grid map is not critical for CBE, and other techniques such as \textit{Gaussian process occupancy maps} (GPOM) \citep{OCallaghan2012} or Hilbert maps \citep{ramos2015hilbert} can be used. However, working with a different type of occupancy map requires changing the calculation of the reward estimation discussed in Section \ref{CBO_exploration}.
  
\subsection{Exploration as an Optimisation Problem}
An optimal path is a series of control inputs $\boldsymbol{u^*}=\vect{u}$ that minimises a desired objective function, $f$, over a valid region $\mathcal{C}$ 
\begin{equation}\label{general_objective}
   \boldsymbol{u^*}=\argmin_{\boldsymbol{u}} f(\boldsymbol{m},\boldsymbol{u}) ~\text{ s.t. } ~\boldsymbol{u} \in \mathcal{C}.
\end{equation}
In most cases, there is no closed-form expression for $f$ or $\mathcal{C}$. Rather, both are \textit{a-priori} unknown and thus can only be estimated from sparse, expensive-to-evaluate and potentially noisy observations (samples). Accordingly, stochastic models are well suited to represent both $f$ and $\mathcal{C}$. However, using probabilistic models for $f$ and $\mathcal{C}$ forces changes in the formulation of the optimisation problem:
\begin{inparaenum}[(i)]  
\item optimisation is performed on  the expected value of the objective function; 
\item  the constraints are estimated using confidence bounds, $\delta$, indicating there is high probability that the constraint is met
\end{inparaenum} \citep{Brochu2010} , which transform Eq. \ref{general_objective} into:
\begin{equation}\label{general_objective_in_expectation}
      	\boldsymbol{u^*}=\argmin_{\boldsymbol{u}} \mathbb{E}[f(\boldsymbol{m},\boldsymbol{u})] \text{ s.t. } \mathrm{Pr}(\mathcal{C}(\boldsymbol{u})) > 1-\delta.
\end{equation}

However, solving Eq. (\ref{general_objective_in_expectation}) in a continuous action space, $\boldsymbol{u}$, is computationally infeasible. A common approach, used by most information-theoretic methods, to reduce complexity is to search for NBV by optimising in pose space. The path planning process is divided into two separate sub-processes. The goal of the first sub-process is to define a set of discretised view points. Each point is the spatial local extermum of the objective function. The second sub-process plans a path from the current location of the robot to its next observation pose and is determined according to obstacles and robot's kinematic constraints (see for example \citet{Makarenko2002, Kollar2008}). 
Although the discrete view points approach simplifies the optimisation process, the resulting path is suboptimal. The main drawback of this approach is that it only considers a limited set of points and thus disregards the potential gains (or the lack thereof) along the entire path. A less obvious, yet significant disadvantage of discrete optimisation, stems from the fact that the expected cost of the resulting path are not an integral part of the view point selection. Penalty heuristics, e.g. distance to goal point, are typically incorporated in the objective function in order to encode cost, yet their limited form underestimates the real cost imposed by motion and safety constraints. As a simple example, consider the case in which the next view point is located close to the robot but is separated by obstacles. While a valid path to that point might exist, it is less desirable due to excessive cost.  

The following section describes our approach for solving Eq. (\ref{general_objective_in_expectation}) by optimising the 
utility of the entire path while minimising the risk of violating any constraints.

\section{Constrained Bayesian Exploration (CBE)}
\label{sec:CBO}
Finding the best path in a continuous space requires optimising a reward function evaluated for any given trajectory. However, computing such a reward for the entire space of paths is computationally infeasible. Furthermore, obstacles and the robot's kinematic envelope impose 
constraints that might not have a closed form expression or are not known \textit{a-priori}. BO provides a strategy to 
learn the reward function and constraints while searching for the valid extremum.

BO guides the optimisation process. Given a sparse training set, BO builds models of the reward function and constraints using Gaussian processes (GPs). 
With these models, BO identifies promising trajectories that correspond to the optimal path with high probability. Constraints are handled in a statistical manner, with BO balancing risk and rewards.

\begin{figure}[bt]
	
	\centering
	
	\includegraphics[height=0.3\textheight,width=0.4\textwidth]{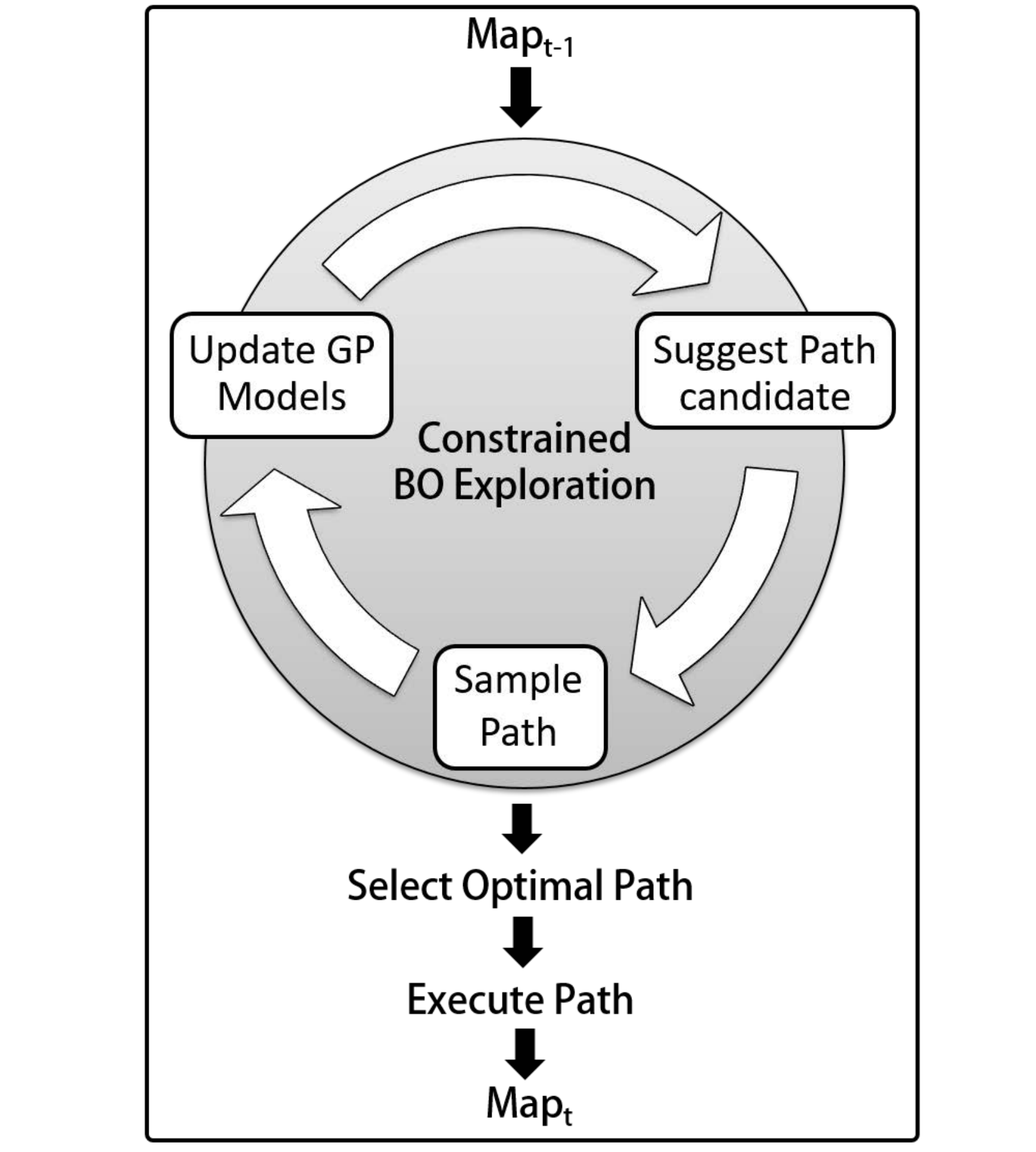}
	
	\caption{A schematic overview of the constrained BO exploration process. Using the current map, BO explores the solution space by guiding the sampling process. Samples are used to update the model belief represented by GPs.  }
	\label{fig:bo_flow_chart}
\end{figure}

A schematic overview of CBE is illustrated in Fig. \ref{fig:bo_flow_chart}. As with any autonomous exploration technique, the expected output of our method is a path that updates our belief about the map. As shown, constrained BO exploration is an iterative process of optimisation and learning. Using the current map as an input, BO explores the solution space by sampling promising path candidates. The samples update the surrogate GP models for the reward and constraints, which is followed by the next BO suggestion. This process continues until resources are exhausted and the optimal path is then selected and executed.

In the following subsections we review the building blocks of CBE. Section \ref{subsec:CBO_GP} provides a short introduction to GP regression and classification, the engine behind the BO surrogate models. Descriptions of the BO method and the constrained BO framework are given in sections \ref{subsec:CBO_BO} and \ref{subsec:CBO_Const_BO}, respectively. In section \ref{CBO_exploration}, the CBE method is described in detail. Finally, a modification of the CBE algorithm to accommodate uncertainty in the robot pose is presented in \ref{subsec:Uncertain_CBE}.

\subsection{Gaussian Processes} \label{subsec:CBO_GP}
\subsubsection{GP regression}
	
GPs are an elegant nonparametric regression technique that places a multivariate Gaussian distribution over the space of 
functions \citep{Rasmussen2006}. A GP is completely defined by a mean function $m(\boldsymbol{x})$ and a covariance or kernel function $k(\boldsymbol{x},\boldsymbol{x}')$ given an input space $\boldsymbol{x} \in \mathbb{R}^D$. Both functions are user-defined and encode prior belief about the properties of the regression problem.
 
A GP builds a model of the unknown underlying function $f(\boldsymbol{x})$ based on a set of noisy observations $\mathcal{D}=\{\boldsymbol{x}_i,y_i\}_{i=1}^N$. 
Here, $\boldsymbol{x}_i \in \mathbb{R}^D$ is the location of the $i$-th sample point and $y_i \in \mathbb{R}$ is the corresponding target value. 
In addition, there is no direct access to $f(\boldsymbol{x})$ and only noisy observations from $f$ are available according to $y=f(\boldsymbol{x})+\epsilon$, where the noise follows a Gaussian distribution $\epsilon \sim \mathcal{N}(0,\sigma_n^2)$. 

Conditioned on observations $\mathcal{D}$, inference at test point $\boldsymbol{x}^\star$ corresponds to the predictive distribution for $y^* = f(\boldsymbol{x^*})$: 	
\begin{equation}
		y^* \mid x^*,\mathcal{D} = \mathcal{N}(\mu,\sigma^2),
\end{equation}
where,
\begin{equation}\label{eq:GP_mu}
			\mu = \boldsymbol{K}(\boldsymbol{x^*},\boldsymbol{X})[\boldsymbol{K}(\boldsymbol{X},\boldsymbol{X})+\sigma_n^2\mathit{I}]^{-1}\boldsymbol{y},
\end{equation}
\begin{multline}\label{eq:GP_sigma}
			\sigma^2 = \boldsymbol{K}(\boldsymbol{x^*},\boldsymbol{x^*})- \\
					\boldsymbol{K}(\boldsymbol{x^*},\boldsymbol{X})[\boldsymbol{K}(\boldsymbol{X},\boldsymbol{X})+\sigma_n^2\mathit{I}]^{-1}k(\boldsymbol{X},\boldsymbol{x^*}).
\end{multline}	
Here $\boldsymbol{X} = \{\boldsymbol{x}_i \in \mathcal{D}\}_{i=1}^N$ and $\boldsymbol{x^*}$ is the test point. $\boldsymbol{K}(\boldsymbol{X},\boldsymbol{X})$ is the covariance matrix between all pairs of inputs in the training set. $\boldsymbol{K}(\boldsymbol{X},\boldsymbol{x^*})$ is the covariance matrix calculated between the test point, $\boldsymbol{x^*}$, and the observation set. Every element in the covariance matrix is calculated from the kernel function $k(\boldsymbol{x},\boldsymbol{x}')$ and its associated set of hyperparameters $\boldsymbol{\theta}$.
	
The choice of hyperparameters, $\boldsymbol{\theta}$, is critical for a successful GP model. The right value for the hyperparameters ensures that the model generalises well without over-fitting the data. The common practice for training a GP's hyperparameters is to maximise the model log marginal likelihood
		\begin{equation}\label{hypoerparam}
			\boldsymbol{\theta}_{optimal} = \argmax_{\boldsymbol{\theta}}\left[log \left(p(y|\boldsymbol{X},\theta \right) \right].    
		\end{equation}
We refer the reader to an extensive discussion on GP hyperparameters training in \citep{Rasmussen2006}.

\subsubsection{GP Least Squares Classifier (GPC)}	

GP Least Squares Classifier (GPC) is a simple and efficient classification method based on GP regression \citep{Rasmussen2006}. 
In binary classification, the observation targets $y_i$ can be either $\{-1,+1\}$ with a corresponding probability of $p$ and $1-p$, respectively. 
With the GPC model, the squared error of a training data point is minimised by mapping the output of a GP regressor to the $[0,1]$ 
interval to obtain the class probability for a test point, $p(y^*|\mathcal{D},\boldsymbol{x}^*)$. This is attained by post-processing the output of the regressor using a "squashing" sigmoid that confines the regression output to the interval $[0,1]$. Using a cumulative Gaussian sigmoid, the class probability is computed as
\begin{equation}\label{GPC}
p(y^*|x^*,\mathcal{D},\boldsymbol{\theta})=\Phi\left(\frac{y^*(\alpha\mu(x^*)+\beta)}{\sqrt{1+\alpha^2\sigma(x^*)^2}}\right),  
\end{equation}
where $y^*$ is either $+1$ or $-1$. $\mu$ and $\sigma$ are calculated from Eqs. (\ref{eq:GP_mu}) and (\ref{eq:GP_sigma}), respectively, $\Phi$ stands for the Gaussian cumulative distribution, $\boldsymbol{\theta}$ represents the hyperparameters of the GP regressor and $\alpha$ and $\beta$ are the "squashing" parameters of the classifier. Training the GPC hyperparameters ($\boldsymbol{\theta}$, $\alpha$, $\beta$) is a two-step process. First, the regression hyperparameters, $\boldsymbol{\theta}$,  are trained by maximising the log marginal likelihood as with conventional 
GP regression. Second, the  squashing parameters, $\alpha$ and $\beta$ are trained by maximising the log predictive probability using a \textit{leave-one-out cross validation} (LOOCV) \citep{Rasmussen2006}: 
\begin{equation}\label{GPC_LOOCV}
L_{LOOCV}=\sum_{i=1}^{N} p(y^i|\boldsymbol{X},y_{-i},\alpha,\beta)  
\end{equation}
The ${-i}$ subscript indicates that target $y_i$ is excluded. The GPC framework provides closed-form expressions for $p(y^i|\boldsymbol{X},y_{-i},\alpha,\beta)$ and for the derivatives of $L_{LOOCV}$ with regards to the hyperparameters, $\alpha$ and $\beta$, which facilitate an efficient training process.  
	
\subsection{Bayesian Optimisation} \label{subsec:CBO_BO}
BO is a powerful global optimiser, which is most effective when 
the objective function does not have a closed-form expression, costly to evaluate and there is no access to derivative information. Given a limited 
set of noisy observations and prior beliefs about the properties of the objective function, BO exploits Bayes' theorem to determine the most 
effective course of action \citep{Brochu2010}. 

The building blocks of a Bayesian optimiser are the surrogate and acquisition functions. The surrogate function 
is the estimated model of the objective function we would like to optimise. It holds our current belief of the underlying function, which is inferred from 
observations and prior knowledge of its properties. Gaussian processes (GPs) are generally used for modelling the 
surrogate function due to their Bayesian non-parametric properties and analytical form. When modelled using a 
Gaussian process (GP), the surrogate function is represented by the posterior mean and variance.

The acquisition function guides the selection of new observation points to sample from the unknown objective function. Based on the current model of the objective function, it provides a quantitative measure for the probability of finding the global extremum in a specific location. The Bayesian optimiser uses this measure as a utility proxy to select the next observation point. In essence, the original optimisation problem, Eq. (\ref{general_objective_in_expectation}), is transformed into an iterative optimisation process of the acquisition function:
\begin{equation}\label{general_acquisition_function_optimisation}
\boldsymbol{x}^{\boldsymbol{u}}_{k+1} = \argmin_{\boldsymbol{x}^{\boldsymbol{u}}}  s(\boldsymbol{x}^{\boldsymbol{u}}).
\end{equation}
The superscript \textsuperscript{$\boldsymbol{u}$} is used for brevity and implies that the result of the optimisation process is a set of control inputs, $\boldsymbol{u}$, as defined in Section \ref{sec:explo}, i.e. $\boldsymbol{x}^{\boldsymbol{u}} =\boldsymbol{u}$.

Although this is still a non-convex optimisation problem, using an appropriate acquisition function makes the search for the extremum more efficient. A proper acquisition function balances the exploration-exploitation trade-off. Therefore, the optimisation process considers both where we believe the extremum lies and promising location in unexplored regions. Consequently, the number of expensive evaluations of the objective function is kept to a minimum. \citet{Brochu2010} list the most commonly used acquisition functions. A modified version of these acquisition functions for autonomous exploration will be introduced in section \ref{subsec:CBO_Const_BO}. Based on the predictive mean, $\mu$, and variance, $\sigma^2$ defined in Eqs. (\ref{eq:GP_mu}) and (\ref{eq:GP_sigma}), the acquisition functions take the following analytic form:
\begin{enumerate}
	\item \textit{Expected Improvement} (EI).
	EI is defined as the expected difference from the true extermum. On its $k+1$-th iteration, the optimiser finds a location that maximises the expected difference from the true extremum, $f(\boldsymbol{x}^{\boldsymbol{u}}_{min})$. In a minimisation problem, finding $f(\boldsymbol{x}^{\boldsymbol{u}}_{min})$, EI is defined as follows: 
	\begin{equation}\label{general_EI_acquisition_function}
		EI(\boldsymbol{x}^{\boldsymbol{u}})= \mathbb{E}\{ f(\boldsymbol{x}^{\boldsymbol{u}}) - f(\boldsymbol{x}^{\boldsymbol{u}}_{min}) \}.
	\end{equation}
	We follow a slightly modified version of EI that uses the predicted mean, $f^*_{min}$, which is inferred from the GP model \citep{Gramacy2011}. Furthermore, we exploit the GP structure to produce a concise closed-form for EI:
	\begin{equation}\label{Expected_Improvement}
		EI(\boldsymbol{x}^{\boldsymbol{u}})=
		\begin{cases}
			-\sigma(\boldsymbol{x}^{\boldsymbol{u}})[Z\Phi(Z)+\phi(Z)] & \text{$\sigma(\boldsymbol{x}^{\boldsymbol{u}}) > 0$}\\
			0 & \text{$\sigma(\boldsymbol{x}^{\boldsymbol{u}})=0$}
		\end{cases}\\
	\end{equation}	
	Where $\phi$ and $\Phi$ represent the normal distribution PDF and CDF, respectively. $\sigma(\boldsymbol{x}^{\boldsymbol{u}})$ is the standard deviation of the posterior distribution in $\boldsymbol{x}^{\boldsymbol{u}}$. $Z$ is given by
    \begin{equation*}
        Z=
	    \begin{cases}
	    (f^*_{min} - \mu(\boldsymbol{x}^{\boldsymbol{u}}) - \zeta)/\sigma(\boldsymbol{x}^{\boldsymbol{u}}) & \text{$\sigma(\boldsymbol{x}^{\boldsymbol{u}}) > 0$}\\
	    0 & \text{$\sigma(\boldsymbol{x}^{\boldsymbol{u}})=0$}
    \end{cases},\\
	\end{equation*}	
	 where $\zeta$ is a user-defined parameter that balances the  exploration-exploitation trade-off.				
	
	\item \textit{Lower Confidence Bound} (LCB).
	The LCB lacks the rigour of EI. However its user-defined parameter, $\kappa$, provides a simple mechanism to adjust the optimisation exploration-exploitation trade-off:
	\begin{equation}\label{LCB}
		LCB(\boldsymbol{x}^{\boldsymbol{u}})=\mu(\boldsymbol{x}^{\boldsymbol{u}})-\kappa\sigma(\boldsymbol{x}^{\boldsymbol{u}})
	\end{equation}
\end{enumerate}

The pseudo code shown in Algorithm \ref{basic_BO_algo} outlines the typical steps performed by BO. In each iteration, a new sampling location, $\boldsymbol{x}^{\boldsymbol{u}}_i$, is found by minimising the acquisition function $s(\boldsymbol{x}^{\boldsymbol{u}})$. BO evaluates the objective function, $f$, at  $\boldsymbol{x}^{\boldsymbol{u}}_i$ and checks whether a new extremum has been found. In addition, the new observation, $f_i$, updates the surrogate (GP) model, which holds our belief of $f$. The updated models are then used on the next iteration of BO. By choosing an appropriate GP model and acquisition function, BO keeps the number of function evaluations low, leading to an efficient optimisation process. 

\begin{figure}[bt]

 \centering
 
 \includegraphics[height=0.3\textheight,width=0.45\textwidth]{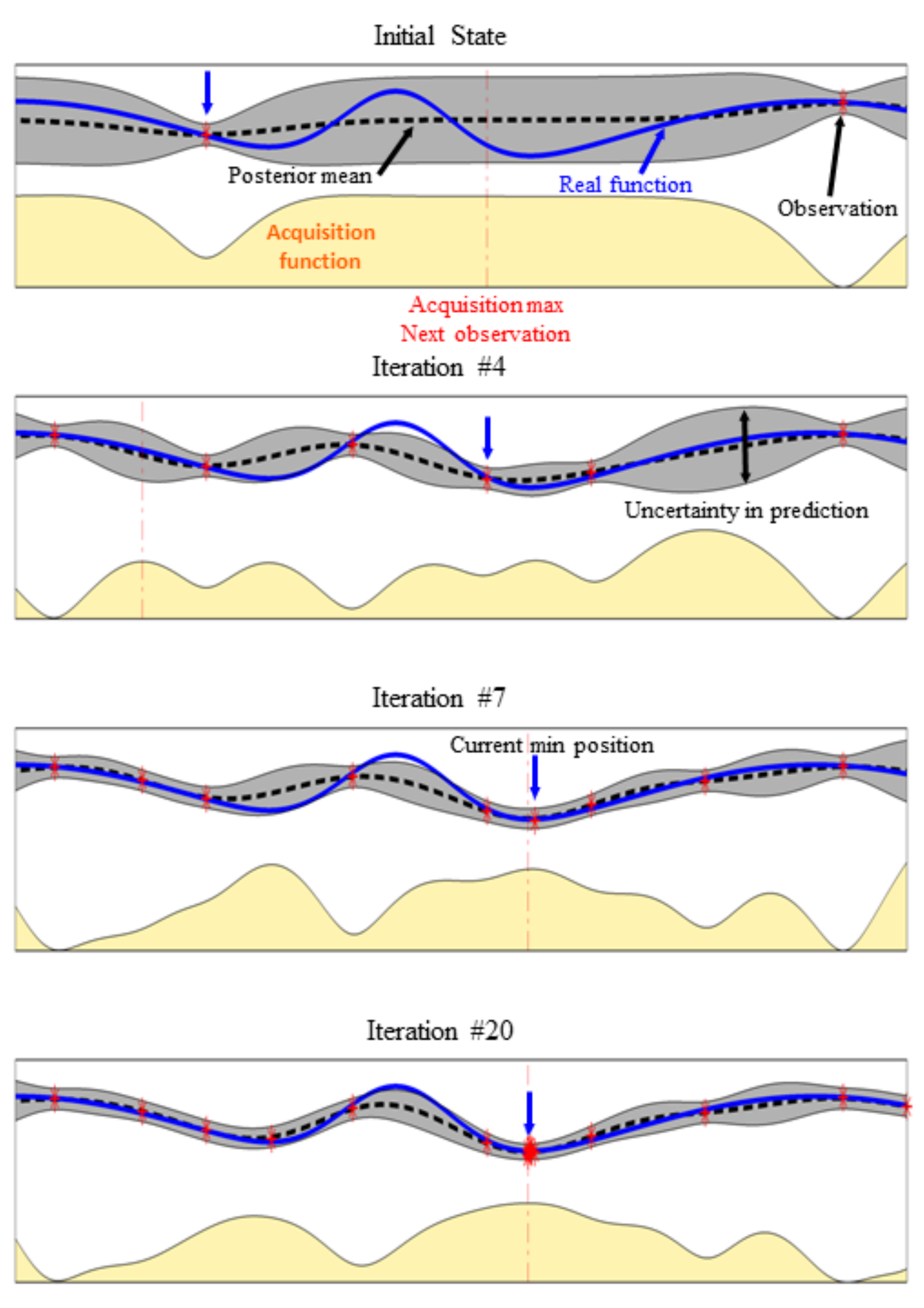}
 
 \caption{One dimensional example of BO. The continuous blue line is the unknown objective function. The red asterisks are samples (with added noise) of this function. The black dashed line and shade represent the posterior GP mean and variance calculated from samples, respectively. The yellow shade is the acquisition function (LCB) which is scaled and with an offset for visualisation purposes. The red vertical dash-dot line represents the next sampling locations, while blue downwards arrow marks the location on the current minimum. }
 \label{basic_bo_fig}
\end{figure}

A one dimensional example of BO is depicted in Fig. \ref{basic_bo_fig}. The optimiser has no knowledge of the objective function (blue line) other than the noisy samples (red asterisks). A GP model is generated based on these observations. The model is accurate and confident around the sampling points, where the posterior mean (black dashed line) converges to the objective function values and its variance (grey shade) is low. Initially, the LCB acquisition function resembles the GP variance, which leads to an aggressive exploratory behaviour at the beginning of the optimisation. As the model becomes more confident, the optimiser focuses its search around the global minimum.

\begin{algorithm}[bt]
	\caption{Bayesian Optimisation}
	\label{basic_BO_algo}
	\DontPrintSemicolon
	\KwIn{${f(\boldsymbol{x}^{\boldsymbol{u}})}$: Objective function.}
	\myinput{${s(\boldsymbol{x}^{\boldsymbol{u}})}$: Acquisition function.}
	\myinput{$f^*_{min}$: Current minimum.}
	\KwOut{$\boldsymbol{x}^{\boldsymbol{u}}_{min}$} 
	\For{$i=1,2,3,...$}{
		Find: $\boldsymbol{x}^{\boldsymbol{u}}_{i} \leftarrow \underset{\boldsymbol{x}^{\boldsymbol{u}}}{\mathrm{argmin}} ~ s(\boldsymbol{x}^{\boldsymbol{u}})$\;
		Sample objective function: $f_i \leftarrow f(\boldsymbol{x}^{\boldsymbol{u}}_i)$\;
		Update GP model with new observation ($\boldsymbol{x}^{\boldsymbol{u}}_{i},f_i$)\;
		\If {$f_i < f^*_{min}$} {
			{$\boldsymbol{x}^{\boldsymbol{u}}_{min} \leftarrow \boldsymbol{x}^{\boldsymbol{u}}_i$} \;
			{$f^*_{min} \leftarrow f_i $}
		}
	}	
\end{algorithm}

\subsection{Constrained Bayesian Optimisation}\label{subsec:CBO_Const_BO}
Constrained BO handles optimisation with unknown constraints. Similarly to the optimisation process discussed in the previous section, our only knowledge of the underlying constraints stems from observations. Furthermore, the objective function is undefined wherever the constraints are violated. To tackle the uncertainty in both objective function and constraints, we employ a constraint weighted acquisition function \citep{Gelbart2014}. Consequently, the optimisation balances the expected reward with the confidence in the constraint model and its associated risk.

In the literature, there are two different modifications to the basic BO acquisitions function relevant to our case. \citet{Gramacy2011} propose a variant of EI called \textit{integrated expected conditional improvement} (IECI). In its essence, IECI represents the marginal effect that a new observation will have on the overall uncertainty of the GP model regardless of its actual value and is defined as
\begin{equation}\label{IECI}
		IECI(\boldsymbol{x}^{\boldsymbol{u}})= \int \mathbb{E}\{EI(\boldsymbol{x}^{\boldsymbol{u'}}|\boldsymbol{x}^{\boldsymbol{u}})\}c(\boldsymbol{x}^{\boldsymbol{u'}})d\boldsymbol{x}^{\boldsymbol{u'}} 
\end{equation}
To incorporate constraints, Gramacy and Lee integrated the conditional improvement under the probability density of the constraint function, $c(\boldsymbol{x}^{\boldsymbol{u'}})$. The main caveat of this method is its scalability. Calculating the integral over the expected conditional improvement requires heavy Monte Carlo sampling of the GP model. Hence, IECI is not a practical method for real-time problems. Furthermore, this method might not be suitable for most standard constrained optimisation problems since it assumes that the objective function can be sampled in regions where the constraint is violated. 

The other modification to BO, which we use in this work, is the constraint weighted acquisition function proposed by \citet{Gelbart2014}. The confidence in the validity of the solution scales the expected utility of the acquisition function. With independent constraints, a Constrained LCB (CLCB) is thus defined as:  
\begin{equation}\label{CLCB_real}
	CLCB(\boldsymbol{x}^{\boldsymbol{u}}) = LCB(\boldsymbol{x}^{\boldsymbol{u}})\prod_{k=1}^{K} \mathrm{Pr}(\mathcal{C}_k(\boldsymbol{x}^{\boldsymbol{u}}) <  1-\delta_k).
\end{equation}
Where $\delta_k$ is a user defined constrained confidence bound over constraint $k$.

In order to model the constraints $\mathcal{C}_k(\boldsymbol{u})$, we employ GPCs, $g_k$, to provide an estimate for the likelihood constraint $k$ is satisfied within the user defined confidence bounds $\delta$:
\begin{equation}\label{CLCB_GPC_real}
	CLCB(\boldsymbol{x}^{\boldsymbol{u}}) = LCB(\boldsymbol{x}^{\boldsymbol{u}})\prod_{k=1}^{K} \mathrm{Pr}(g_k(\boldsymbol{x}^{\boldsymbol{u}}) <  1-\delta_k).
\end{equation}
Incorporating learned constraints complicates the optimisation algorithm as shown in pseudo code in Algorithm \ref{Constrained_BO_algo}. Since the objective function is undefined in regions where the constraints are not satisfied, a preprocessing step finds feasible and valid regions. Within these regions, the optimiser finds the next observation using the utility function defined in Eq. (\ref{CLCB_GPC_real}). With every new observation point, the constraints are assessed and their respective GPC model is updated. The GP model for the goal function, on the other hand, is only updated when all constraints are met.

\begin{figure}[bt]
	
	\centering
	
	\includegraphics[height=0.3\textheight,width=0.45\textwidth]{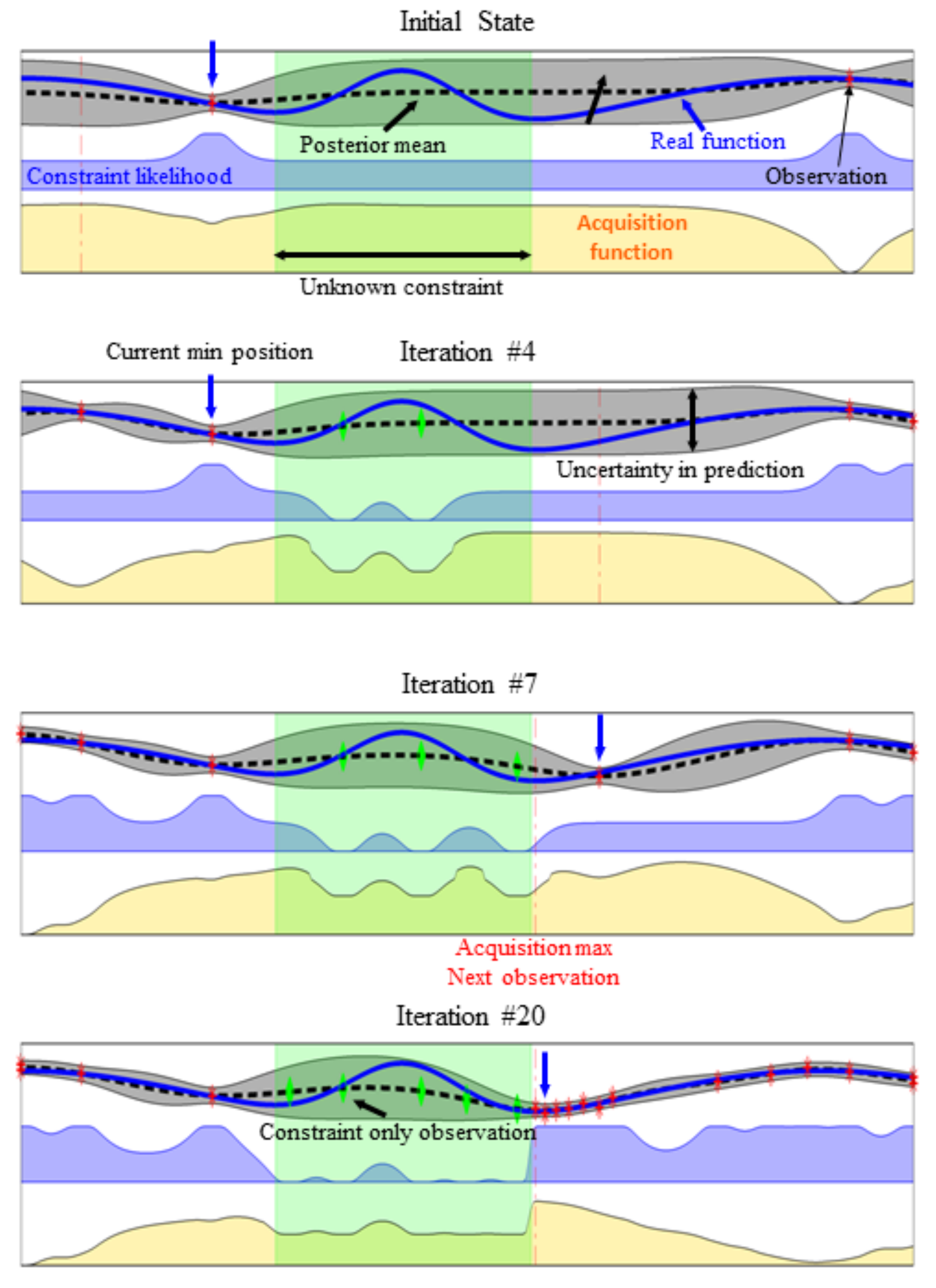}
	
	\caption{One dimensional example of BO with an unknown constraint. The continuous blue line is the unknown goal function and the green area indicates the unknown constraint. The green diamonds are observations of violated constraint, hence the goal function can not be sampled. The red asterisks are samples (with added noise) of the goal function, where the constraint is met. The black dashed line and shade represent the posterior GP mean and variance calculated from samples, respectively. The area in blue is the constraint likelihood function, where high values stand for high probability that the constraint will be satisfied. The yellow area is the basic unconstrained acquisition function (LCB). Both acquisition and constraint likelihood functions are scaled and with an offset for visualisation purposes. The red vertical dash-dot line represents the next sampling locations, while blue downwards arrow marks the location on the current minimum}
	\label{constrained_bo_fig}
\end{figure}

A one dimensional example for constrained BO is shown in Fig. \ref{constrained_bo_fig}. The unknown constraint is indicated by the area shaded in green, while its predictive probability is represented by the blue area. As with the regression of the objective function, the confidence in the constraints value, whether valid or invalid, is higher around observation points. As evident from Fig. \ref{constrained_bo_fig}, BO tries to evaluate points outside the constrained region, however this only updates the constraint model while the objective GP model is unchanged (hence uncertainty is high). With every observation, BO becomes more confident in the model of the objective function, borders of the constraint and the location of the global minimum.
\begin{algorithm}[bt]
	\caption{Constrained Bayesian Optimisation}
	\label{Constrained_BO_algo}
	\DontPrintSemicolon
	\KwIn{${f(\boldsymbol{x}^{\boldsymbol{u}})}$: Objective function.}
	\myinput{${s(\boldsymbol{x}^{\boldsymbol{u}})}$: Acquisition function.}
	\myinput{${g_k(\boldsymbol{x}^{\boldsymbol{u}})}$: $k$-th constraint function.}
    \myinput{$\delta_k$: $k$-th constraint tolerance.}
	\myinput{$f^*_{min}$: Current minimum.}
	\KwOut{$\boldsymbol{x}^{\boldsymbol{u}}_{min}$} 
	\For{$i=1,2,3,...$}{
		feasible region $\mathcal{C}$: $\mathcal{C}(\boldsymbol{x}^{\boldsymbol{u}}) = \prod_{k=1}^{K} \mathrm{Pr}(g_k(\boldsymbol{x}^{\boldsymbol{u}}) <  1-\delta_k)$\;
		Find next sample point: \;
		 $\boldsymbol{x}^{\boldsymbol{u}}_{i} \leftarrow \underset{\boldsymbol{x}^{\boldsymbol{u}} \in \mathcal{C}}{\mathrm{argmin}} ~ s(\boldsymbol{x}^{\boldsymbol{u}})\prod_{k=1}^{K} \mathrm{Pr}(g_k(\boldsymbol{x}^{\boldsymbol{u}}) <  1-\delta_k)$\;
		\If {$\boldsymbol{x}^{\boldsymbol{u}}_{i}$ valid}{
			Sample objective function: $f_i \leftarrow f(\boldsymbol{x}^{\boldsymbol{u}}_i)$\; 
			Update GP model with new observation ($\boldsymbol{x}^{\boldsymbol{u}}_{i},f_i$)\;
			If {$f_i < f^*_{min}$}: {$\boldsymbol{x}^{\boldsymbol{u}}_{min} \leftarrow \boldsymbol{x}^{\boldsymbol{u}}_i$}
			}
		Update GPCs with new observation\;
	}	
\end{algorithm}

\subsection{Constrained Bayesian Exploration (CBE)}\label{CBO_exploration}
When planning a path for autonomous exploration, the objective is to acquire new knowledge and improve the existing map in an efficient manner. Yet, both robot and environment impose restrictions on the solution space. Any selected path should be safe, i.e. free of obstacles, and within the kinematic envelope of the robot. Constrained BO is a very flexible tool to find such a solution. Safety and motion restrictions are treated as constraints, which are learnt while the optimiser searches for the global extremum among the possible paths.    

CBE is an iterative process, of finding a promising sampling location, observing the unknown objective and constraint function values at that location before updating the model using the observed values. In the context of robotic exploration, sampling a location corresponds to testing a new path candidate. It is important to note that while CBE optimises the controls $\boldsymbol{u} \in \mathbb{R}^D$, the objective function and constraints are observed along trajectories, $\xi(\boldsymbol{u}) \in \mathcal{T}$. While $\mathcal{T}$ is a subspace of the robot's configuration space, we avoid the explicit notation, since the exact structure of the configuration space is unknown. The pseudo code for assessing a new action, i.e. new path, is shown in Algorithm \ref{path_check} and it categorically consists of two major parts:
\begin{inparaenum}[(i)]  
	\item Path Candidate Validity Assessment,
	\item Reward calculation. 
\end{inparaenum}

\subsubsection{Path Candidate Validity Assessment}\hfill \break
  	The role of the path candidate validity assessment is to determine whether a path is valid or not. A valid path is safe from collisions with obstacles and within the kinematic capabilities of the robot. 
	
	There are two tests to assess the validity of a new path. First, the maximum curvature, $\kappa_{max}$, along the path is evaluated. If it exceeds a user-defined threshold, $\kappa_{max} > \delta_{\kappa}$, then the path is considered invalid and no other tests are needed. Although this is a relativity simple constraint, it has the  potential to incorporate other motion considerations, such as energy or execution time budgets. Furthermore, learning the motion constraints provides greater flexibility when responding to changing driving conditions.
	
	The second test validates the safety of a path. Given the occupancy map, it identifies obstacles along the path and dead-ends. Additionally, to ensure safety, the planned path should not traverse unobserved parts of the map. Formally, we require that the occupancy along a safe path should not exceed a user-defined confidence threshold, $\delta_{safe}$:
	\begin{equation}\label{eq:safety}
		Safe(\boldsymbol{u}) = 
		\begin{cases}
		1 & \text{ $\boldsymbol{m}[\xi(\boldsymbol{u})]\ < \delta_{safe}$}\\
		0 & o.w.\\
		\end{cases}
	\end{equation}
	It is important to note again, that the result of Eq. (\ref{eq:safety}) is used as a point observation in the generation of a stochastic model for the safety constraints. As this is a learned model, which is based solely on these observations, the exact implementation of the safety criteria is unimportant. Instead of using $\boldsymbol{m}[\xi(\boldsymbol{u})]\ < \delta_{safe}$, the user can define a different test to assess path safety that better suits the robot and environment configuration. The simplest example changes the confidence safety threshold, $\delta_{safe}$, which will modify the risk-reward balance. More complicated methods may use different occupancy maps or include visibility test for dynamic obstacles.
	      
	In case the path is invalid, an additional post-processing step is taken in order to better define the valid solution space. The invalid path is expanded into a subset of derived valid and invalid paths, which are used to update the GP models. The added observations help reduce the uncertainty of the relatively sparse GP models. But more importantly, it provides the GP model with the boundaries between the valid and invalid space.
	
\subsubsection{Reward calculation}	\hfill \break
	In autonomous exploration, a commonly optimised objective function is \textit{information gain} (IG). Information gain measures the reduction in the map entropy after observations are made
	\begin{equation}\label{IG}
		IG(\boldsymbol{u}) = H(\boldsymbol{m}) - H(\boldsymbol{m}|\boldsymbol{u})
	\end{equation}
	Here $ H(\boldsymbol{m})$ is the entropy of the map and $H(\boldsymbol{m}|\boldsymbol{u})$ is the entropy after a path, defined by a set of actions $\boldsymbol{u}$, was executed. In order to evaluate the information gain along an entire path, it is necessary to emulate the expected laser observations along the path and their accumulated effect on the map's statistical model. This is achieved by ray tracing the simulated beams originating from a robot moving on the path.
	
	For an effective optimisation, we limit the action search space to be a parametrised trajectory originating from the robot's current pose. In this work we use quadratic splines, however other forms of trajectory representation can be easily utilised. The limitation of parametrised trajectories is their expressivity, especially when constraints are present. The effect this limitation has on the optimisation process are more profound in areas that were explored by the robot before or around dead-ends. In such areas, the reward is similar in all valid paths. Meaning that with the limited choice of paths and under the imposed constraints, the robot can not resolve a viable decision that will pull it toward unexplored regions. To alleviate this problem, we introduce two penalties that affect the optimisation process only when the local differences are small:
	\begin{itemize}
		\item The first penalty provides a global context to the overall objective function. A coarse path is planned from the robot's location to the nearest frontier. This path does not have to be traversable by the robot and it can violate safety or kinematic envelope constraints. However, it biases path selection toward a region where IG reasoning is feasible. We define a penalty value, $P_{H}(\boldsymbol{u})$, which is a function of the difference between the direction of the assessed path and the coarse path. Therefore, a path that develops in the opposite direction of the global coarse path will have higher penalty than a path that is oriented towards a similar direction.
		
		\item The second penalty is a function of the path length, $P_{L}$. The rationale is to penalise very short and longs paths. This will drive the robot forward, whilst preventing overly confident longer paths.
	\end{itemize}

    The additional penalties are added to the IG reward with corresponding weights, $W_1$ and $W_2$. These weights keep the penalties small compared to the typical IG utility:
   	\begin{equation}\label{ModifiedIG}
    		IG_{Modified}(\boldsymbol{u}) = 	IG(\boldsymbol{u}) + W_1\cdot P_{H}(\boldsymbol{u}) + W_2\cdot 	P_{L}(\boldsymbol{u})
   	\end{equation}
    
	Even with the simplest occupancy map representation, the forward projection model needed to estimate IG is expensive to evaluate. This is the main motivation for using BO; optimising decision making while keeping sampling low. Instead of optimising by explicitly calculating the forward simulation IG results, BO learns IG by sparsely sampling it and building an equivalent model. It then uses these models to infer the next sampling location. The efficiency of BO relies on the accuracy of the learned GP models. However, a high fidelity GP surrogate model requires a substantial number of function observations. To increase the number of sample points and fill in the gaps in the GP model, we notice that the IG along the path is a non-decreasing monotonic function. Since the robot motion along the path is a set of sequential observation points, the IG in any given point is the sum of accumulated IG of all previous observations and the contribution of the current observation
	\begin{equation*}
		IG([u_1...u_{k+1}]) = IG([u_1...u_{k}])+\delta IG(u_{k+1} | z_{k+1})
	\end{equation*}
	Thus, by evaluating the path reward sequentially, a denser GP model for the objective function can be generated at no additional computational cost.

\begin{algorithm}[bt]
	\caption{CBE Path assessment}
	\label{path_check}
	\KwIn{$\xi(\boldsymbol{u})$: assessed path}
	\myinput{$f_{min}$: current objective minimum}
	\KwOut{$\boldsymbol{P}_{valid}$, $\boldsymbol{P}_{reward}$, $f_{min}$}
	\
	\DontPrintSemicolon
	
	$\boldsymbol{P}_{valid} \leftarrow$ Check: Motion Constraints\;
	$\boldsymbol{P}_{valid} \leftarrow$ Check: Safety\;
	\eIf{$\boldsymbol{P}_{valid}$}{
		$\boldsymbol{P}_{reward}\leftarrow$ Evaluate reward: eq. \ref{ModifiedIG};\\
		If $\boldsymbol{P}_{reward}$ $< f_{min}$: {$f_{min}$=$\boldsymbol{P}_{reward}$}
	}{
		Path assessment(valid subset of $\boldsymbol{P}$)\;
	}

\end{algorithm}
\begin{algorithm}[bt]
	\caption{CBE}
	\label{Safe_Autonomous_Exploration_algo}
	\DontPrintSemicolon
	
	/* Generate initial training set: */\;
	$N=$ Size of training set \;
	$\Omega \leftarrow$ Generate training path set(N)\;
	\For{$\boldsymbol{u}_{k} \in \Omega$}{
		
		$\boldsymbol{u}_{k}\leftarrow$ Path assessment($\boldsymbol{u}_{k}$) (Algorithm \ref{path_check}) \;
		Update GP and GPCs: $\boldsymbol{u}_{k}$\;
	}
	
	/* Constrained BO: */\;
	\For{$i=1,2,3,...$}{
		feasible region  $\mathcal{C}$: $\mathcal{C}(\boldsymbol{u}) = \prod_{k=1}^{K} \mathrm{Pr}(\mathcal{C}_k(\boldsymbol{u}) <  1-\delta_k)$\;
		Find: $\boldsymbol{u}_{i} \leftarrow \underset{\boldsymbol{u} \in \mathcal{C}}{\mathrm{argmin}} ~ C(\boldsymbol{u})\cdot LCB(\boldsymbol{u})$\;
		$\boldsymbol{u}_{i}\leftarrow$Path assessment($\boldsymbol{u}_{i}$) (Algorithm \ref{path_check})\;
		Update GP and GPCs: $\boldsymbol{u}_{i}$\;
	}
	
	Execute	optimal path\;
\end{algorithm}	
\begin{figure}[]
	
	\centering
	\captionsetup[subfigure]{justification=centering}
	
	\subfloat[Generating Training set]{\includegraphics[height=0.3\textheight,width=0.45\textwidth]{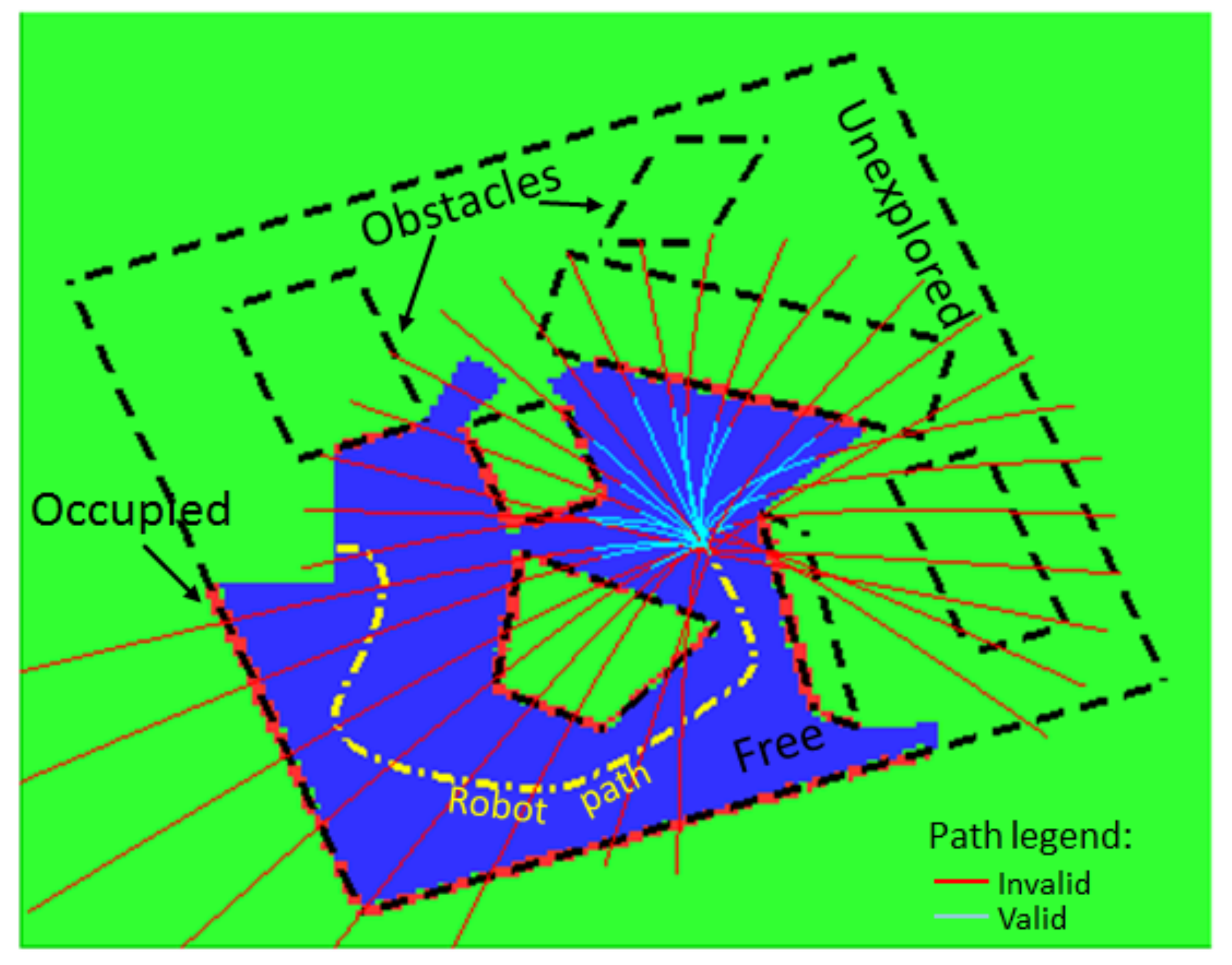}
		\label{fig:exp_Generating_Training_set}}
	
	\subfloat[Optimisation]{\includegraphics[height=0.3\textheight,width=0.45\textwidth]{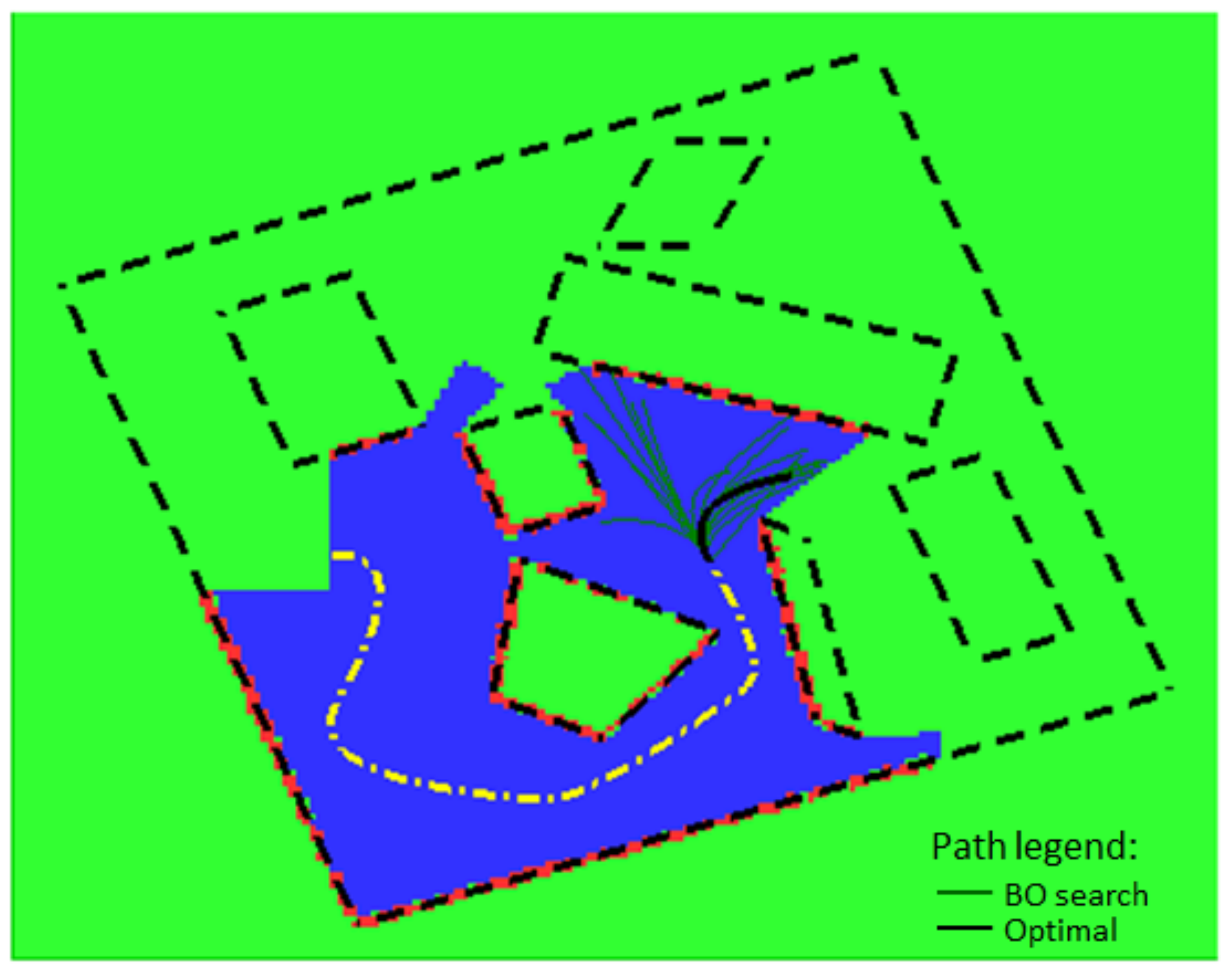}
		\label{fig:exp_Optimisation}}
	\caption{CBE searches for an optimal path in an unexplored room. Walls and obstacles are denoted by the dashed black lines, which maybe unknown to the robot. Green areas are unexplored, blue are known to be free while red areas contain known obstacles. \protect\subref{fig:exp_Generating_Training_set} The paths used in the training set are shown in red (invalid) and cyan (valid). \protect\subref{fig:exp_Optimisation} The CBE optimisation process produces optimal path candidates (green) with the final output of the optimiser sh0own in black.}
	\label{Spline_explanation}
\end{figure}

\begin{figure*}[htbp]
	
	\centering
	
	\includegraphics[width=0.68\textwidth]{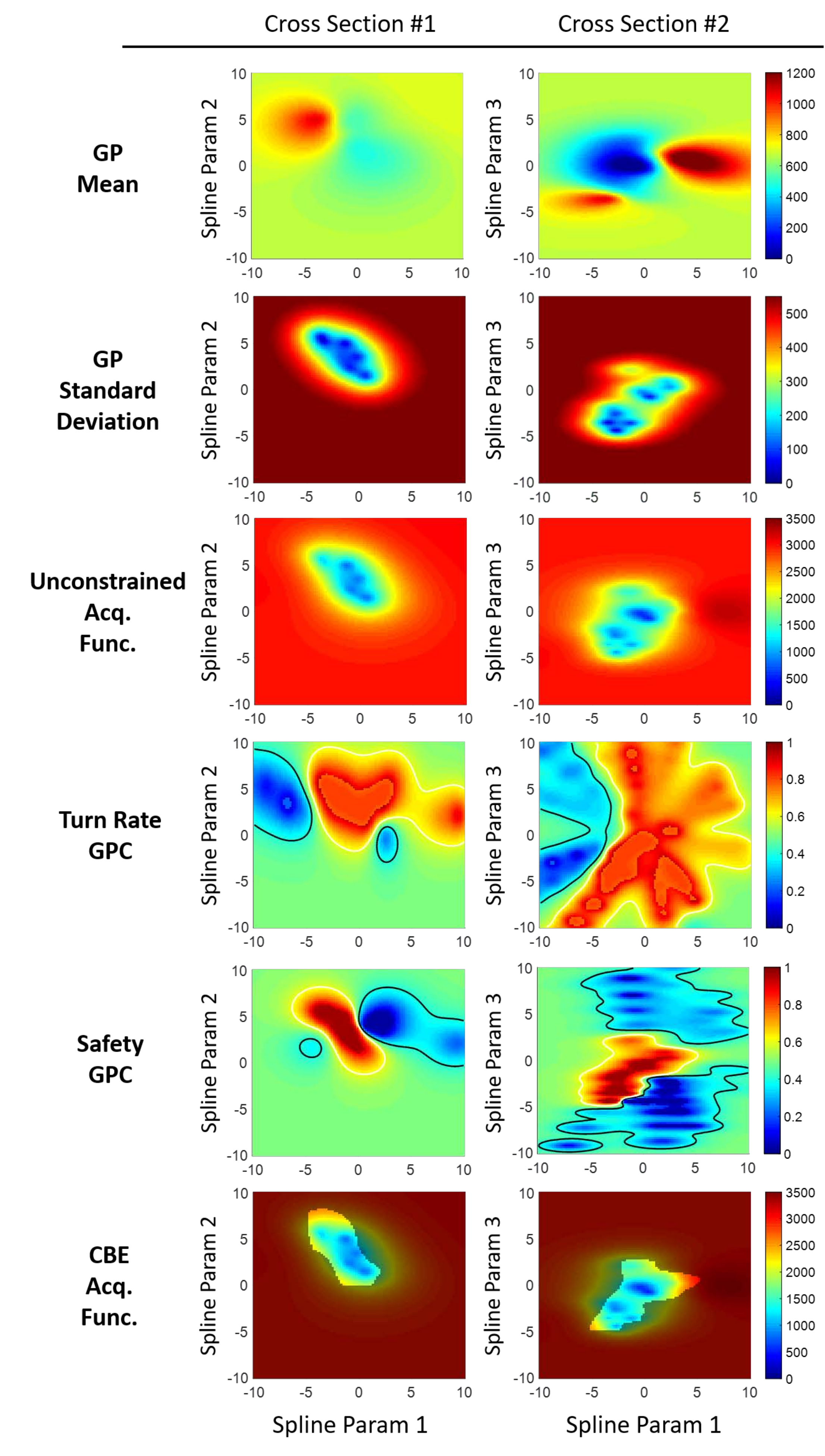}
	
	\caption{Images (cross sectional) of the various components of CBE for the scenario shown in Fig. \ref{Spline_explanation}. Rows depict different component of BO; GP regression mean, GP standard deviation, unconstrained BO acquisition function, turn rate GPC, safety GPC and the CBE acquisition function. Columns show two orthogonal cross sections. Contours in GPC images represent valid (white) and invalid (black) thresholds. Black contours for the CBE acquisition function (last row) define valid regions for optimisation given observations. CBE will try to maximise acquisition function in valid regions, which produces a suggestion for the next observation point.    }
	\label{fig:Spline_BO_example}
\end{figure*}

\subsubsection{CBE Algorithm}	\hfill \break

A pseudo code for CBE using constrained Bayesian optimisation is given in Algorithm \ref{Safe_Autonomous_Exploration_algo}. Fig. \ref{Spline_explanation} provides a visual explanation of this process. The algorithm is divided into two parts. First, a training set consisting of a small number of paths, typically 20-50, is formed. The training set is shown in Fig. \ref{fig:exp_Generating_Training_set} by the red (invalid) and cyan (valid) paths. As explained before, the algorithm tries to extract valid path segment from an invalid path as can be seen in Fig. \ref{fig:exp_Generating_Training_set}. These paths, valid as well as invalid, are used in the update of the GP and GPCs models, which serve as a prior model for the subsequent constrained BO stage. 

The second part of Algorithm \ref{Safe_Autonomous_Exploration_algo} is the constrained BO. In Fig. \ref{fig:exp_Optimisation}, the outputs of this stage, which correspond to the paths suggested by the optimiser, are depicted in dark green. With every attempt, BO updates the GP and GPCs and becomes more confident in the model of the objective function, the constraints and the location of the global minimum. This learning process is evident from the distribution of the suggested paths. Although most are bundled around two main directions, there are some stray paths that check potentially rewarding options. Also, some paths are on the borders of the unexplored regions, suggesting the optimiser tries to learn about the motion constraints. The final output of the optimiser, the optimal path where the accumulated reward is maximised, is shown in black.

To gain additional insight into the optimisation process, Fig. \ref{fig:Spline_BO_example} presents images of key CBE elements. As CBE is a high-dimensional optimisation process, cross-sections are used for their visualisation. The first key element of CBE is the surrogate GP model of the objective function, shown in the first two cross sections, $\mu_{GP}$ and $\sigma_{GP}$. The GP represents our belief about the learned objective function. The main benefit of using a GP, as with other Bayesian regression techniques, is the ability to obtain an inference confidence measure. The non-parametric structure gives great flexibility in expressing the model expected value and variance around observations. Previous method using BO for exploration, e.g. in  \citep{Martinez-Cantin2009}, optimises using the unconstrained acquisition function which is shown inverted in Fig. \ref{fig:Spline_BO_example} for clarity. Instead of only optimising over the expected value, the use of an acquisition function incorporates the model uncertainty. However, unconstrained BO is not suitable for autonomous exploration, as reward samples can only be acquired along valid trajectories. Beyond the valid region, the GP model provides only its intrinsic parameters; the model mean and maximum variance. Consequently, this breaks the internal BO feedback loop of sampling and updating, as the GP model is kept unchanged after any invalid sample. 

The constrained BO framework is more suitable for autonomous exploration. The learned constraints, shown as GPCs in Fig. \ref{fig:Spline_BO_example}, provide the optimiser with an additional layer that incorporates invalid samples without the need to define a closed-form expression for the constraints. Using GPCs provides an efficient method to query the certainty in which the constraints are met. Furthermore, the user can easily modify the validity threshold (shown as black and white lines), to adjust the optimiser risk-reward balance. The combined acquisition function, which is shown in the last cross-section of Fig. \ref{fig:Spline_BO_example}, is simply the unconstrained BO acquisition function overlaid with the GPCs valid zone. Unlike unconstrained BO, integrating new invalid samples will not break the BO feedback loop. It will instead modify the GPCs. This behaviour allow the CBE to explore promising paths that are on the borders between valid and invalid.

Finally, we conclude this section with an estimate of the computational complexity of CBE. As GPs are used extensively through out this algorithm, it is not surprising that the CBE computational costs are mainly associated with GP inference complexity. Similarly to other non-parametric methods, the computational complexity depends on the size of training set, $n$. In the CBE framework, however, two separate training sets are defined; 
\begin{inparaenum}[(i)]  
	\item $N$ includes the entire training set of valid and invalid points.
	\item $m$ ($m<N$) includes only the valid points used by the reward GP.
\end{inparaenum}
The computational complexity of a typical GP is $\mathcal{O}(n^3)$ and is due to the Cholesky decomposition of the covariance matrix \citep{Rasmussen2006}. GP prediction carries a lower complexity of $\mathcal{O}(n^2)$ arising from the solution of a triangular linear system. In CBE, on the other hand, we employ $c\sim \mathcal{O}(1)$ GPCs. Hence the overall complexity of the Cholesky decomposition of the various components of CBE is $\mathcal{O}(m^3+cN^3)$. In addition, during optimisation, GP and GPCs model may be queried repeatedly leading to $\mathcal{O}(Mm^2+cPN^2)$, where $P$ and $M$ are the number of queries of GPCs and GP, respectively, and $M<P$. 
Given $m<N$ the overall complexity of CBE, Cholesky decomposition and model queries, can be estimated as $\mathcal{O}(PN^3+cPN^2)$. We can further simplify this expression, by noting that the typical training set contains several hundred points, as is the number of optimisation steps; $P \sim N$. Thus, we can concisely write the overall CBE complexity as $\mathcal{O}(cN^3)$.

\subsection{Incorporating Uncertainty in CBE}
\label{subsec:Uncertain_CBE}

In the context of autonomous exploration, path planning is a decision making process aimed at improving the map fidelity. Any uncertainty, whether it is in sensor observations or in the robot pose, propagates into our belief over the map and corrupts it. While sensor uncertainty is typically a fixed limitation arising from the system configuration, the robot location uncertainty is controlled by the robot's decisions. Reducing pose uncertainty is commonly addressed in the literature by incorporating a "loop closing" heuristic in the optimisation of the next observation point (for example \citep{Makarenko2002,Indelman2015,Rafieisakhaei2016}). With the standard BO framework, incorporating such a heuristic requires modifications to the forward simulation reward calculation as described in the work of \citet{Martinez-Cantin2009}. However, uncertainty in the robot pose necessitates some additional adaptations to the CBE algorithm to ensure the safety of resulting path. Therefore, we will leave the "loop closing" reward modification for future work, and discuss the required changes to CBE in the following section.  

\begin{figure}[bthp]
	
	\centering
	
	\includegraphics[height=0.3\textheight,width=0.45\textwidth]{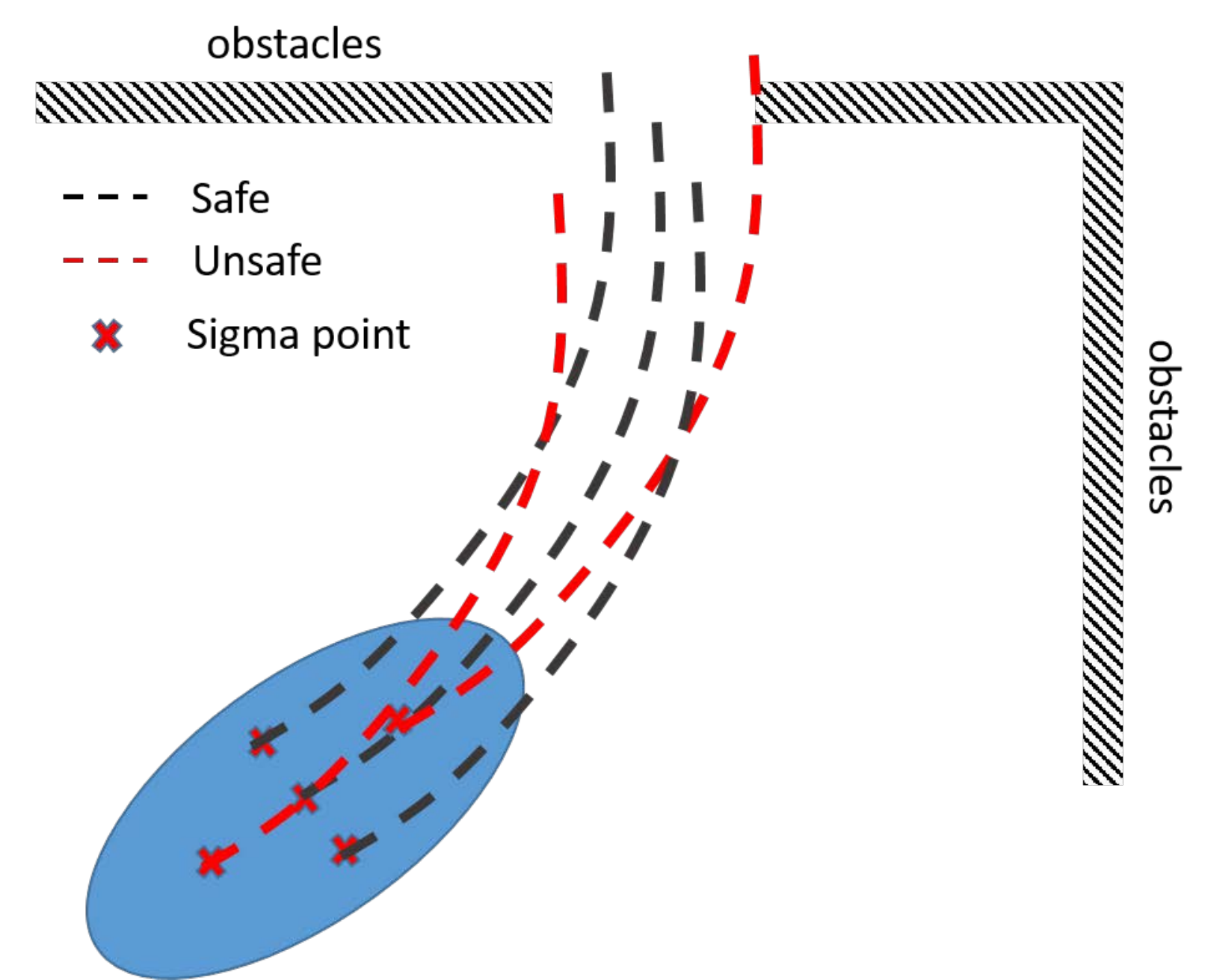}
	
	\caption{Uncertainty in path execution due to uncertain location and orientation results in a non-trivial distribution of the overall path safety.}
	\label{fig:sigma_paths}
\end{figure}

At the end of optimisation, CBE returns an optimal path. When considering only the nominal pose, the optimal path is safe and valid. However, the actual outcome of that path, and more importantly, its safety, depends on the real pose of the robot. Fig. \ref{fig:sigma_paths} depicts an example of such a case, by plotting the same path for several starting poses, drawn from the robot's state distribution. It is clear, that by not incorporating the pose uncertainty, the risk of collision is greatly under-estimated. 

To better estimate the safety risk, we need to project the variance of the robot's location and orientation into the safety GPC model. However, the resulting probability density function might have a non-trivial form. An efficient solution to alleviate this problem utilises an \textit{unscented transform} (UT) \citep{Julier1997}. UT employs a deterministic sampling schedule to estimate the mean and variance of the desired distribution. The sample set, termed 'sigma points', consists of $2n+1$ samples and weights for a $n$-dimensional space. Given the pose mean, $\boldsymbol{\mu_p}$, and covariance, $\boldsymbol{\Sigma_{pp}}$, the 'sigma' points, $\boldsymbol{\chi}$, are defined by the following equations:
\begin{equation}\label{eq:sigma_point}
	\chi_i = 
	\begin{cases}
	\boldsymbol{\mu_p} & i=0\\
	\boldsymbol{\mu_p} + (\sqrt{(n+\lambda)\Sigma_{p}})_i & i=1,...,n\\
	\boldsymbol{\mu_p} - (\sqrt{(n+\lambda)\Sigma_{p}})_{i-n} & i=n+1,...,2n\\
	\end{cases}
\end{equation}
Here $\lambda = \vartheta^2(n+\nu)-n$. $\vartheta$ determines the spread of the sigma points around the mean, $\mu_p$, and is typically a small positive number (in our experiment $\vartheta = 10^{-3}$). $\nu$ is a second order term to adjust kurtosis and is usually set to zero \citep{Wan2000}.

For the sigma weights, we follow \citet{Wan2000}, and define separate values for mean and covariance calculations:
\begin{equation}\label{eq:sigma_w}
\begin{aligned}
& w^{mean}_0 = \frac{\lambda}{n+\lambda} &\\
& w^{cov}_0 = \frac{\lambda}{n+\lambda} + (1-\psi^2+\varrho) &\\
& w^{mean}_i = w^{cov}_i = \frac{1}{2(n+\lambda)} \qquad i=1,...,2\delta&\\
\end{aligned}
\end{equation}
The parameter $\varrho$ is used to encode prior knowledge of initial distribution. Since we assume a Gaussian distribution for the pose, we take $\varrho=2$. 

These sigma points serve as starting poses for alternative path outcomes as shown in Fig. \ref{fig:sigma_paths}. We employ the same mapping from actions to trajectory space, $\xi(\boldsymbol{u})$, but replace the implicit noiseless pose with the sigma points $\xi(\boldsymbol{u},\chi_i)$. As a result, we probe how $\xi$ changes the shape of the initial pose uncertainty and thus recover a stochastic estimate for the robot pose along the path as, $\boldsymbol{x} \sim \mathcal{N}(\rho,\Sigma_{x})$:
\begin{equation}\label{eq:sigma_pred_mean}
\begin{aligned}
& \rho = \sum_{i=0}^{2n} w^{mean}_i\xi(\boldsymbol{u},\chi_i) &\\
& \Sigma_{x} =  \sum_{i=0}^{2n} w^{cov}_i(\xi(\boldsymbol{u},\chi_i)-\rho)(\xi(\boldsymbol{u},\chi_i)-\rho)^T &\\
\end{aligned}
\end{equation}
 
Given $\xi(\boldsymbol{u},\chi_i)$ and map, $\boldsymbol{m}$, one can now stochastically reason about the safety of an action, $\boldsymbol{u}$. Although straight forward, Algorithm \ref{CBE_path_check_uncertain} simplifies the safety estimation even further. Instead of inferring the pose distribution along path, each sigma path, originating from a specific sigma point, is validated separately. The overall validity of an action, $\boldsymbol{u}$, is then determined by the worst-case scenario over all paths. 
 
\begin{algorithm}[bt]
	\caption{CBE Path assessment assuming partially observable pose}
	\label{CBE_path_check_uncertain}
	\KwIn{$\xi(\boldsymbol{u})$: assessed path}
	\myinput{$f_{min}$: current objective minimum}
	\myinput{$\chi$: sigma points}
	\KwOut{$\boldsymbol{P}_{valid}$, $\boldsymbol{P}_{reward}$, $f_{min}$}
	\
	\DontPrintSemicolon
	
	\ForEach{$p \in \chi$}{
	$\boldsymbol{P}_{valid} \leftarrow$ Check: Motion Constraints($p$)\;
	$\boldsymbol{P}_{valid} \leftarrow$ Check: Safety($p$)\;
	}
	\eIf{$\boldsymbol{P}_{valid}$}{
		$\boldsymbol{P}_{reward}\leftarrow$ Evaluate reward: Eq. \ref{ModifiedIG};\\
		If $\boldsymbol{P}_{reward}$ $< f_{min}$: {$f_{min}$=$\boldsymbol{P}_{reward}$}
	}{
	Path assessment(valid subset of $\boldsymbol{P}$)\;
}
\end{algorithm}
 
\section{Results}
\label{sec:experimental}
In this section, we evaluate the performance of CBE in simulation and with a real robot. 

\subsection{Simulations}
We divide our simulation experiments into two categories. First, the robot maps various randomly generated cluttered unstructured environments. The second experiment involves exploration of a large scale complex networks of city roads. In both cases, we simulate a ground robot equipped with a $180^\circ$ field of view (FOV) laser scanner driving at a constant speed. The turn rate of the robot is limited, forcing the optimiser to plan within the robot's kinematic envelope using quadratic splines. In all simulations, we assume full knowledge of the robot's pose.

\subsubsection{Unstructured Environments} 	
\begin{figure}[bt]
 	
 	\centering
 	
 	\includegraphics[width=0.45\textwidth,height=0.5\textheight]{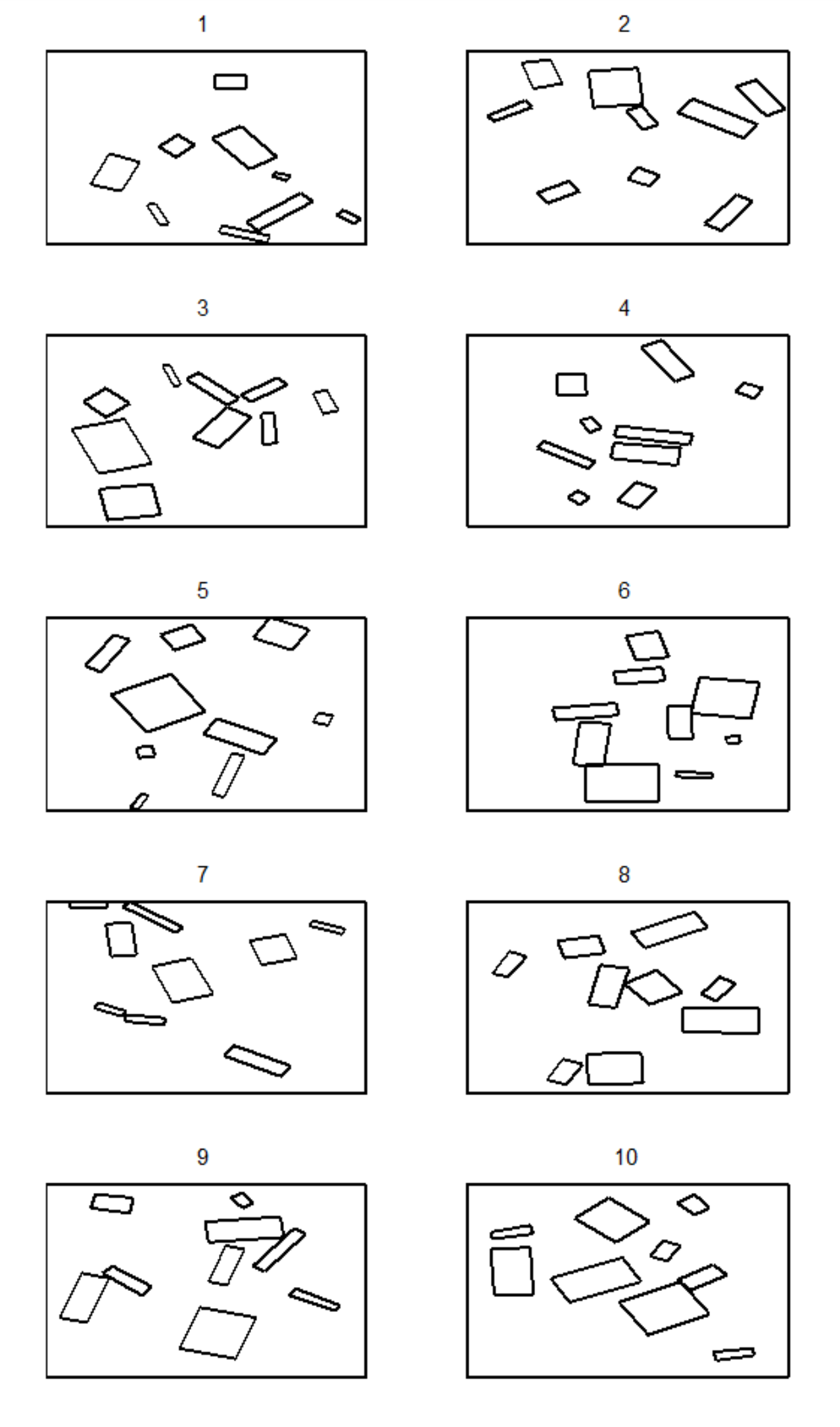}
 	
 	\caption{A sample of randomly generated unstructured worlds used for comparison of exploration methods shown in Table \ref{ExplorationTable}.}
 	\label{RandomWorld}
 \end{figure}

\hfill \break
Many exploration experiments involve structured man-made scenarios. A structured environment is constructed of a network of corridors, for example, underground mines and buildings. Although the unexplored regions include obstacles, the corridor-like structure pulls the robot toward an obvious general path. Unstructured scenes, with randomly positioned obstacles, break any large-scale formation, and thus lack this implicit guidance. Furthermore, these scenarios exhibit additional difficulties, such as isolated areas with only a single access point, nontraversable narrow gaps, and long barriers dividing the world into several almost independent parts. Such an arrangement complicates the exploration process, since it introduces many more options the robot has to choose from. Fig. \ref{RandomWorld} shows examples of randomly generated worlds used to compare constrained BO exploration to other techniques. 

 \begin{figure*}[t]
 	
 	\centering
 	
 	\includegraphics[width=1\textwidth]{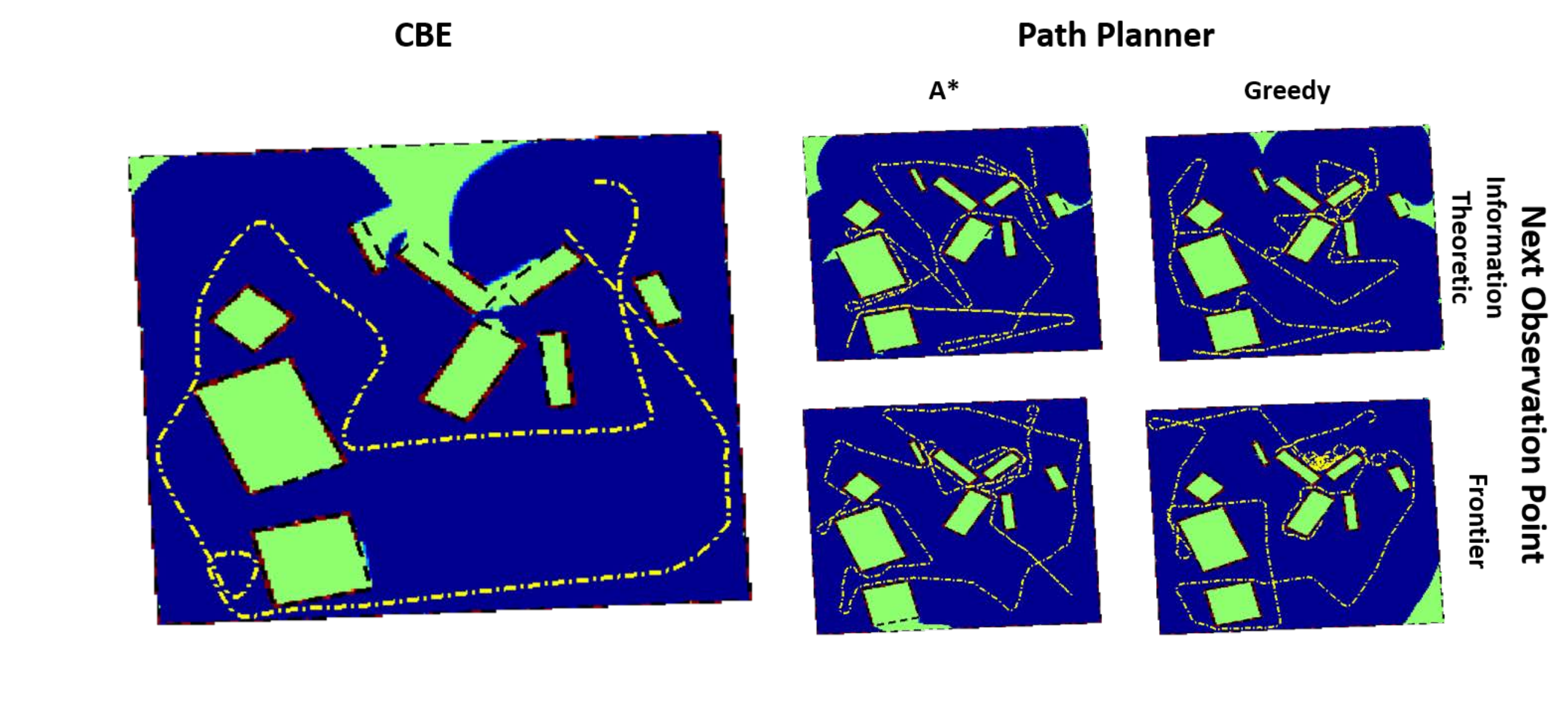}
 	
 	\caption{Comparison of simulation results for a randomly generated world (world 3).(left) CBE results, (right) simulation results with two planners for next observation point; Information-theoretic \citep{Makarenko2002} and frontier \citep{Yamauchi1997}, and two smooth safe path planners, greedy and A*. The walls and obstacles are marked with black lines. In the grid map, green is unexplored regions, blue is free space and red occupied. The executed paths are shown as dashed yellow lines. CBE maximises the accumulated information gain at every decision point by avoiding previously traversed paths. In contrast, both information-theoretic and frontier based planners exhibit a clear criss-cross pattern in the executed paths, as both methods only reason on the gains of single goal point.}
 	\label{PathPlanningComparison}
 \end{figure*}

A qualitative comparison between CBE and other map building techniques is shown in Fig. \ref{PathPlanningComparison}. The methods used for comparison follow the common exploration paradigm, where the path is determined in two separate stages: 
\begin{enumerate}[i]
	\item Selecting the next observation point.
	
	We employ two method for the selection of the next observation point;  a frontier-based method \citep{Yamauchi1997} and an information-theoretic approach based the information gain utility of \citep{Makarenko2002}.
	
	\item Planning a safe path to that goal point.
	
	To emphasise the importance of the path, and not only of the end goal point, we employ two separate path planning techniques for stage two; $A^*$ planner which finds the shortest traversable path to the goal point and a fast greedy planner using the distance to the goal point as its heuristic. Both path planners enforce the robot's safety and manoeuvrability limitations, by generating a path from a valid set of motion primitives.   

\end{enumerate}

The main advantage of CBE visible in Fig. \ref{PathPlanningComparison}, is that the number of overlapping paths is smaller when compared to the other planners. The BO planner takes a relatively short path that minimises the time the robot moves through already visited parts of the environment. CBE maximises the accumulated information gain at every decision point by avoiding previously traversed paths, which is achieved by choosing paths without explicitly defining an end goal point. In contrast to CBE, both information-theoretic and frontier based planners exhibit a clear criss-cross pattern in the executed paths, regardless of the path planner used. This suboptimal performance arises from the two stage exploration process. Choosing a goal point first and then planning a path, prevents reasoning on the potential reward along the driven path. As a result, the knowledge gained while travelling to the goal point is not considered in the decision making.

\begin{figure}[bt]
 	
 	\centering
 	
 	\includegraphics[width=0.45\textwidth]{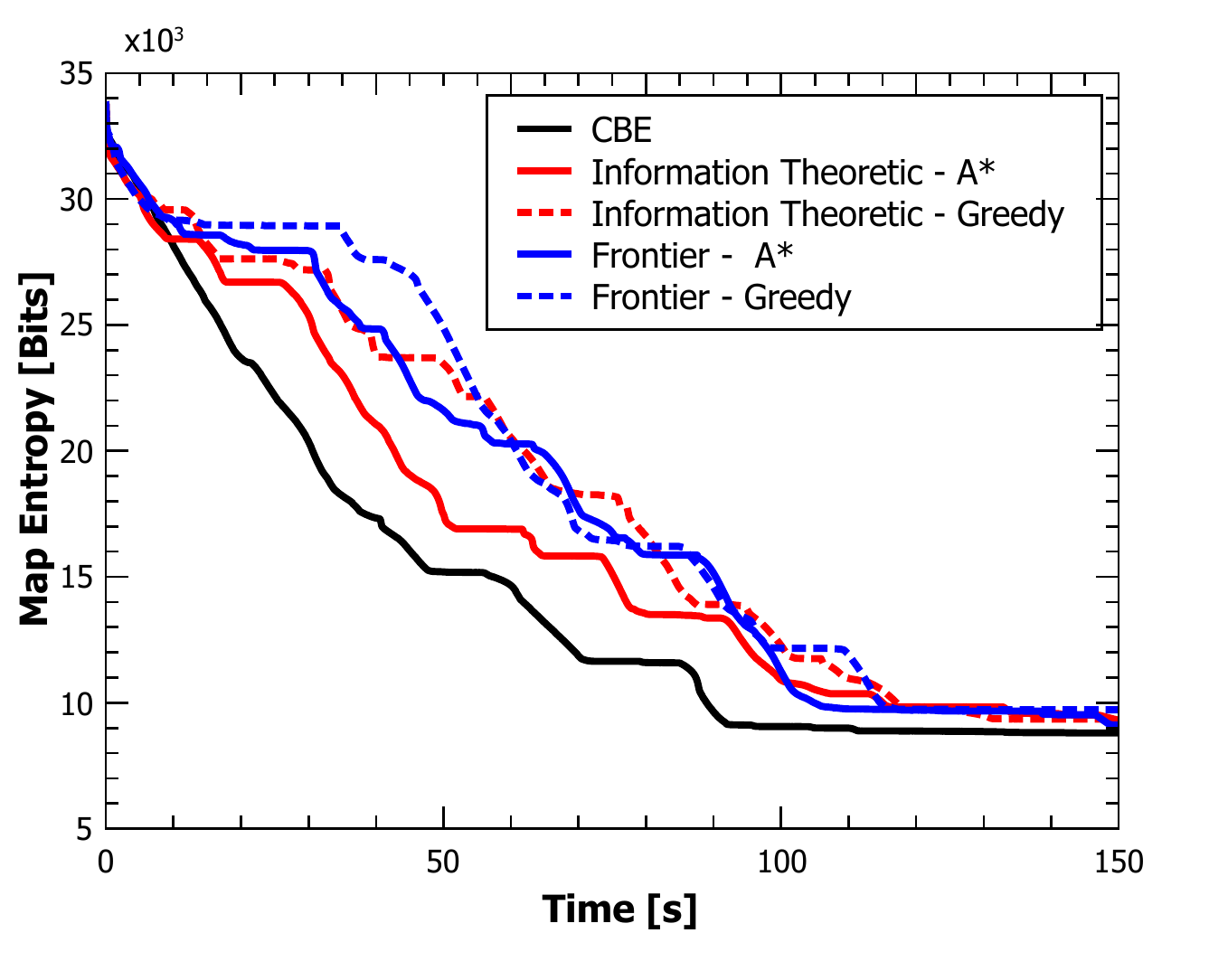}
 	
 	\caption{Quantitative comparison of the reduction in map entropy (world 3) between exploration methods presented in Fig. \ref{PathPlanningComparison}. The initial rate of entropy reduction is similar in all methods. However, the rate slows in both the information-theoretic and frontier methods, as the robot travels through already explored region in route to its next goal point. CBE, avoids previously traversed areas of the map, leading to a faster reduction in entropy. }
 	\label{Comparison_graph}
 \end{figure}

The type of path planner used has also great impact on the overall exploration performance. As expected, the greedy path planner is less effective at finding a path through the clutter, evident by the tangled paths around obstacles. This leads to longer paths with an overall lower rate of improvement. Fig. \ref{Comparison_graph} provides a quantitative comparison of the rate of reduction in the map's entropy between the various methods. The initial rate, in the first 10 seconds, is similar in all methods as the robot passes through the unexplored map. However, the rate at which the entropy decreases in the frontier and information gain based methods becomes slower, coinciding with the robot travelling through already explored regions on its way to the planner's goal point. This outcome is independent of the path planner used,  $A^*$ or greedy. The difference in performance stems from the objective of both the information-theoretic and frontier planners; to find the next global observation point. With a complex unstructured scenario, there are many potential observation points at every decision. It is clear that by visiting these points, the entire environment will be mapped. However, with global point planners there are no guarantees on the optimality of that process, as there is little reasoning about the executed path. While these techniques might put a cost or penalty on the driving distance, the gains along a path are not taken into consideration. CBE, on the other hand, plans in its local neighbourhood taking into account the benefits and risks of potential paths, rather than selecting goal points. The global component pulling the robot toward the nearest frontier only affects the decision when the local information component is negligible. 
\begin{table*}[t]
	\centering
	\caption{Comparison of exploration time between CBE and two planners for next observation point; Information-theoretic \citep{Makarenko2002} and frontier \citep{Yamauchi1997}, and two smooth safe path planners, greedy and A* 
	}
	\label{ExplorationTable}
	\renewcommand{\arraystretch}{1.5}
	\begin{tabular}{|l|c|c|l|l|l|}
		\hline
		& \multicolumn{5}{c|}{Exploration Time {[}s{]}} 
		\\ \hline
		Observation point selection & \multirow{2}{*}{CBE} & \multicolumn{2}{c|}{Information-theoretic}                    & \multicolumn{2}{c|}{Frontier}                            \\ \cline{1-1} \cline{3-6} 
		Path Planner                &                     & Greedy        & \multicolumn{1}{c|}{$A^*$} & \multicolumn{1}{c|}{Greedy} & \multicolumn{1}{c|}{$A^*$} \\ \hline
		& \multicolumn{5}{c|}{}                                                                                                       \\ \hline
		World 1                     & 63.8 (46\%)         & 136.1 (97\%)  & 68.9 (49\%)                & 139.9 (100\%)               & 85.1 (61\%)                 \\ \hline
		World 2                     & 86.2 (64\%)         & 134.5 (100\%) & 82.7 (61\%)                & 87.2 (65\%)                 & 88.5 (66\%)                 \\ \hline
		World 3                     & 86.5 (78\%)         & 108.1 (97\%)  & 98.4 (88\%)                & 111.6 (100\%)               & 99.9 (89\%)                 \\ \hline
		World 4                     & 67.4 (58\%)         & 106.0 (92\%)  & 64.5 (56\%)                & 115.6 (100\%)               & 98.1 (85\%)                 \\ \hline
		World 5                     & 79.6 (89\%)         & 84.4 (94\%)   & 89.8 (100\%)               & 73.5 (82\%)                 & 78.3 (87\%)                 \\ \hline
		World 6                     & 79.4 (17\%)         & 454.6 (100\%) & 76.9 (17\%)                & 120.2 (26\%)                & 80.2 (18\%)                 \\ \hline
		World 7                     & 62.6 (63\%)         & 87.9 (89\%)   & 65.3 (66\%)                & 99.0 (100\%)                & 86.6 (87\%)                 \\ \hline
		World 8                     & 77.1 (52\%)         & 112.4 (76\%)  & 88.7 (60\%)                & 147.4 (100\%)               & 90.7 (62\%)                 \\ \hline
		World 9                     & 71.7 (54\%)         & 129.0 (97\%)  & 115.0 (87\%)               & 132.4(100\%)                & 102.3 (77\%)                \\ \hline
		World 10                    & 76.6 (66\%)         & 115.4(99\%)   & 98.0 (85\%)                & 111.0 (96\%)                & 95.1 (82\%)                 \\ \hline
	\end{tabular}
\end{table*}

Table \ref{ExplorationTable} provides a quantitative comparison between the exploration techniques on several randomly-generated worlds shown in Fig. \ref{RandomWorld}. As expected, $A^*$ is a noticeably better path planner than the greedy planner, as it guarantees the shortest traversable path to the goal point. However, it is not immediately clear, which observation point selection method performs best. Although, in some scenarios the performance of both information-theoretic and frontier methods is similar, there are cases where one method outperforms the other by a significant margin. The CBE method, on the other hand, consistently maintains good performance. In the majority of the tested scenarios, CBE is the fastest method. In all other cases, it has similar performance as the leading method, whether frontier or information-theoretic. These results show that the CBE planner is less sensitive to the layout of the environment and provides a more consistent and robust method for exploration compared to the other techniques.

\begin{figure}[bthp]
	
	\centering
	
	\includegraphics[width=0.49\textwidth]{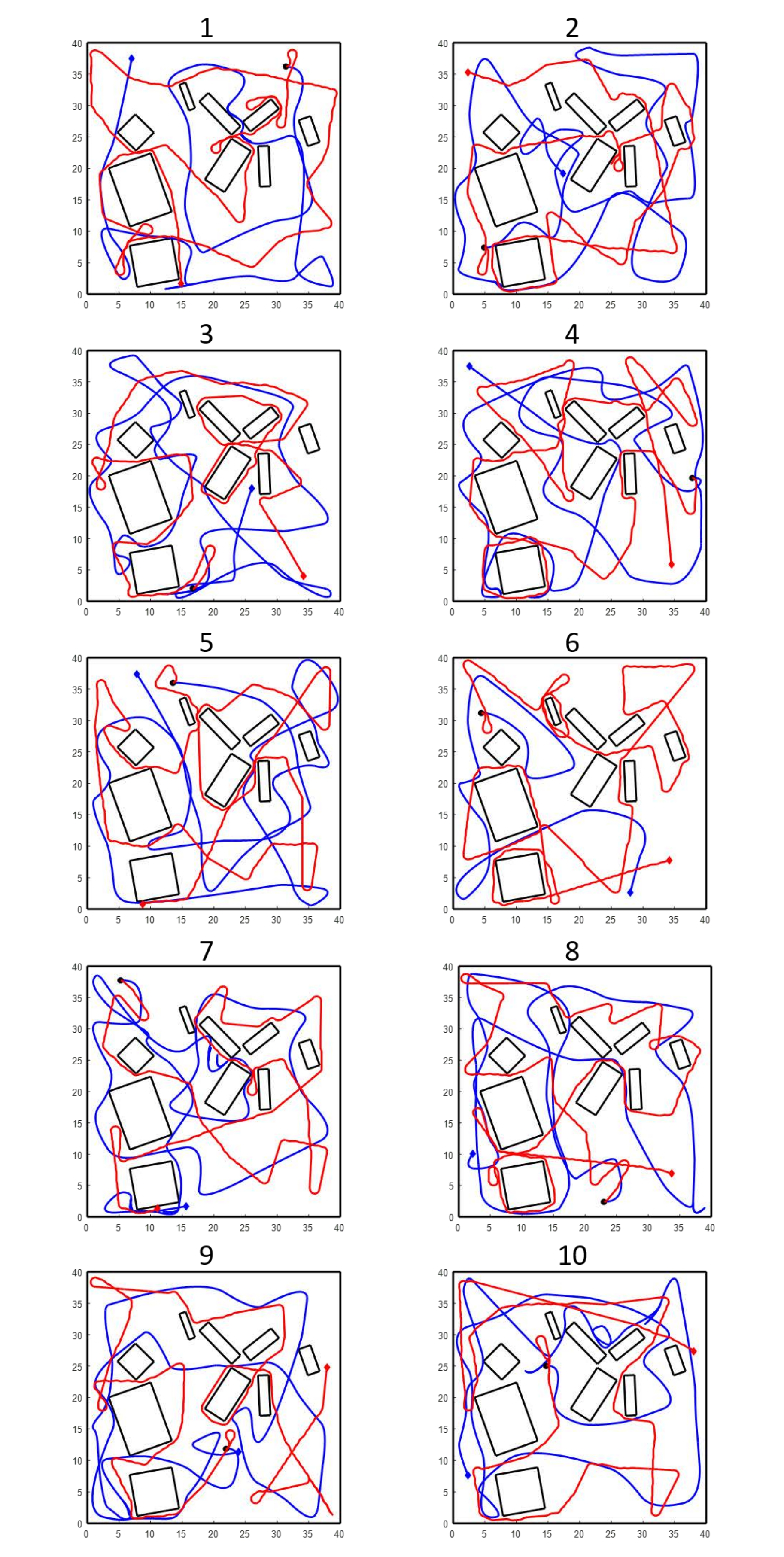}
	
	\caption{Repeatability tests. Each image depicts exploration paths in world 3 (refer to Fig. \ref{RandomWorld}) from different starting poses, marked by a black dot. The blue and red lines are the exploration paths, CBE and frontier \citep{Yamauchi1997}, receptively. Each path ends with a corresponding diamond shaped marker.} 
	\label{fig:world3_exploration_comparison}
\end{figure}

As we assume the robot pose is fully known in these simulations, repeatability was tested by changing the initial pose. Fig. \ref{fig:world3_exploration_comparison} and Table \ref{tab:World3_comparison} present a comparison between CBE and frontier/$A^*$ in world 3 (see in Figs. \ref{RandomWorld} and \ref{PathPlanningComparison}) for various starting poses. Once again, we can see that CBE outperforms the competing method in most cases.   

\begin{table}[]
	\centering
	\caption{Repeatability -  quantitative comparison of exploration paths originating from various starting poses as shown in Fig. \ref{fig:world3_exploration_comparison}. Comparison is between CBE and frontier \citep{Yamauchi1997} exploration methods.}
	\label{tab:World3_comparison}
	\begin{tabular}{|c|c|c|c|}
		\hline
		\multirow{2}{*}{\#} & \multicolumn{2}{c|}{Exploration Time {[}s{]}} & \multirow{2}{*}{Diff {[}\%{]}} \\
	   \cline{2-3} 
		& CBE                 & Frontier               &               \\
		\hline
		1                   & 128.9               & 107.3                  & -16.8         \\
		\hline
		2                   & 89.8                & 111.9                  & 24.5          \\
		\hline
		3                   & 68.7                & 104.6                  & 52.2          \\
		\hline
		4                   & 96.9                & 103.9                  & 7.3           \\
		\hline
		5                   & 72.7                & 97.4                   & 34.1          \\
		\hline
		6                   & 61.1                & 78.7                   & 28.8          \\
		\hline
		7                   & 61.8                & 98.6                   & 59.7          \\
		\hline
		8                   & 88.4                & 91.4                   & 3.4           \\
		\hline
		9                   & 99.5                & 98.5                   & -1.0          \\
		\hline
		10                  & 70.0                & 76.9                   & 9.8           \\
		\hline
		\multicolumn{4}{|c|}{ } \\
		\hline
		Average             & 83.8                & 96.9                   & 20.2
		\\
		\hline
		std. dev.             & 21.2                & 11.6                   & 24.1
		\\
		\hline             
	\end{tabular}
\end{table}

\subsubsection{Structured Environments} 	\hfill \break

These experiments test CBE performance in a structured environment scenario. Part of the roads and paths network of Venice and Jerusalem old city were extracted from Google Maps. These complex networks of corridor-like patterns serve as the ground truth in this large-scale exploration experiment. In such a structured system, there is no clear advantage for the constrained BO method. The corridor structure forms an obvious path, which limits the local significance of path selection. As there are little differences in rewards along the paths, the end goal becomes the most important property of a path. Hence, a CBE planner would be potentially ineffective. By comparison, the frontier based-$A^*$ approach seems to be the most sensible method for such a problem, as it moves the robot on the shortest path to the edge of the known space.
  \begin{figure}[]
  	
  	\centering
  	
  	\includegraphics[width=0.45\textwidth,height=0.79\textheight]{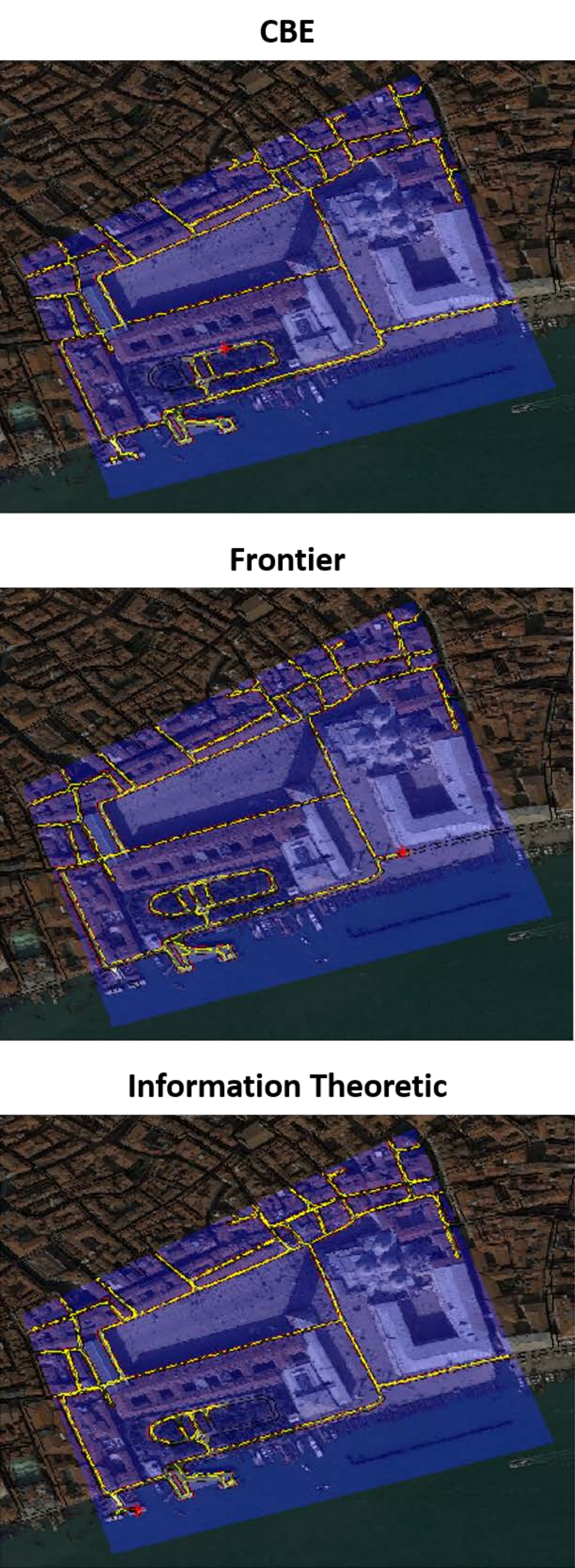}
  	
  	\caption{Comparison of simulation results of map building in Venice. Top to bottom; CBE, frontier and information-theoretic based exploration. Traversable roads are extracted from Google Maps. The blue background marks the mission area. Explored areas are shown without the blue background. The path the robot executed is in yellow, and the last position of the robot is marked with a red asterisk. As expected, all methods explored almost the entire mission area. The paths generated by the information-theoretic method revisits explored region of the map much more than the other methods, although a distance penalty is incorporated in its reward heuristic. CBE and frontier present similar path structure as both plan mostly in the robot's close neighbourhood.}
  	\label{VeniceMapComparison}
  \end{figure}

  \begin{figure*}[bt]
  	
  	\centering
  	
  	\includegraphics[width=1\textwidth]{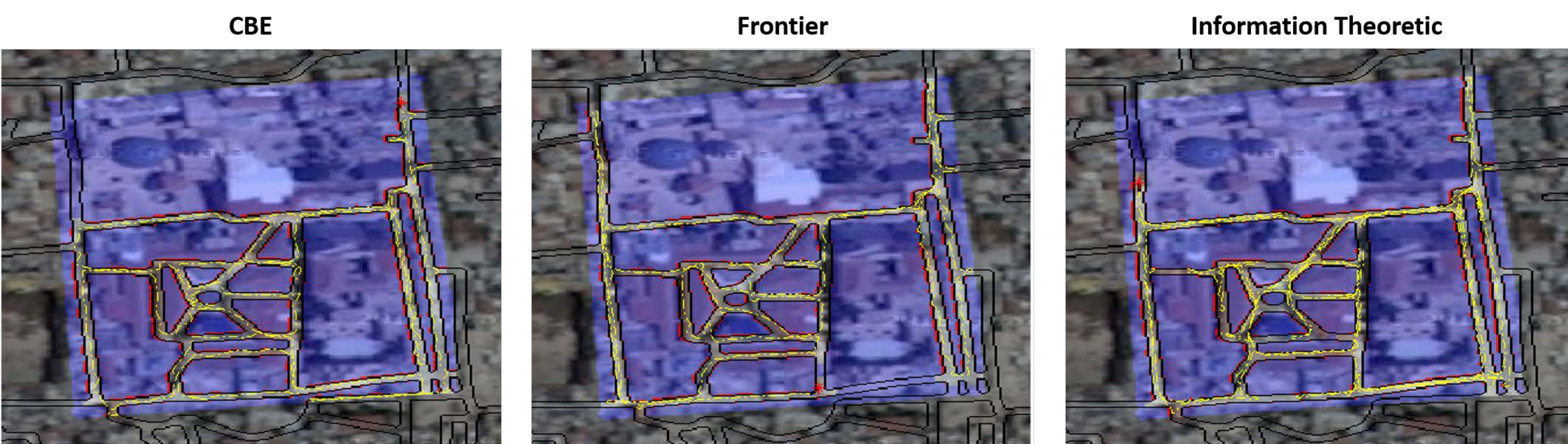}
  	
  	\caption{Comparison of simulation results of map building in Jerusalem old city. Left to right; CBE, frontier and information-theoretic. Traversable roads are extracted from Google Maps. The blue background marks the allowed region. Explored area are shown without the blue background. The path the robot executed is  in yellow, and the last position of the robot is marked with a red asterisk.  As expected, all methods explored almost the entire mission area. The paths generated by the information-theoretic method revisits explored region of the map much more than the other methods, although a distance penalty is incorporated in its reward heuristic. CBE and frontier present similar path structure as both plan mostly in the robot's close neighbourhood.}
  	\label{JeruslaemMapComparison}
  \end{figure*}
    
  \begin{figure}[]
  	
  	\centering
  	
  	\includegraphics[width=0.45\textwidth]{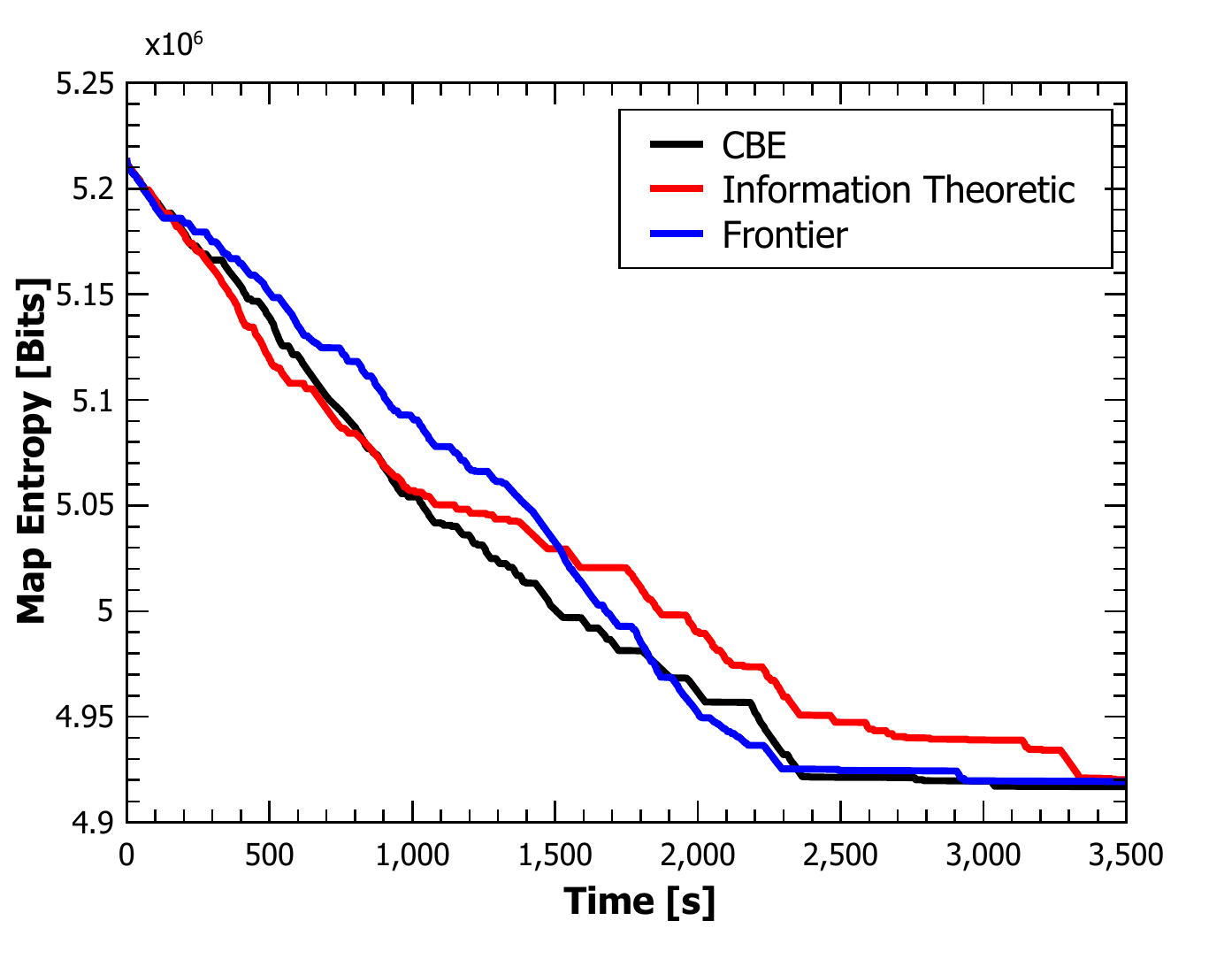}
  	
  	\caption{Venice - Comparison of reduction in map entropy between exploration methods presented in Fig. \ref{VeniceMapComparison}. The overall time to cover the mission area is similar with both CBE and frontier. Both methods outperforms the information-theoretic method as the number of paths crossing already explored regions of the map is lower.}
  	\label{VeniceComparisonGraph}
  \end{figure}
  
  \begin{figure}[]
  	
  	\centering
  	
  	\includegraphics[width=0.45\textwidth]{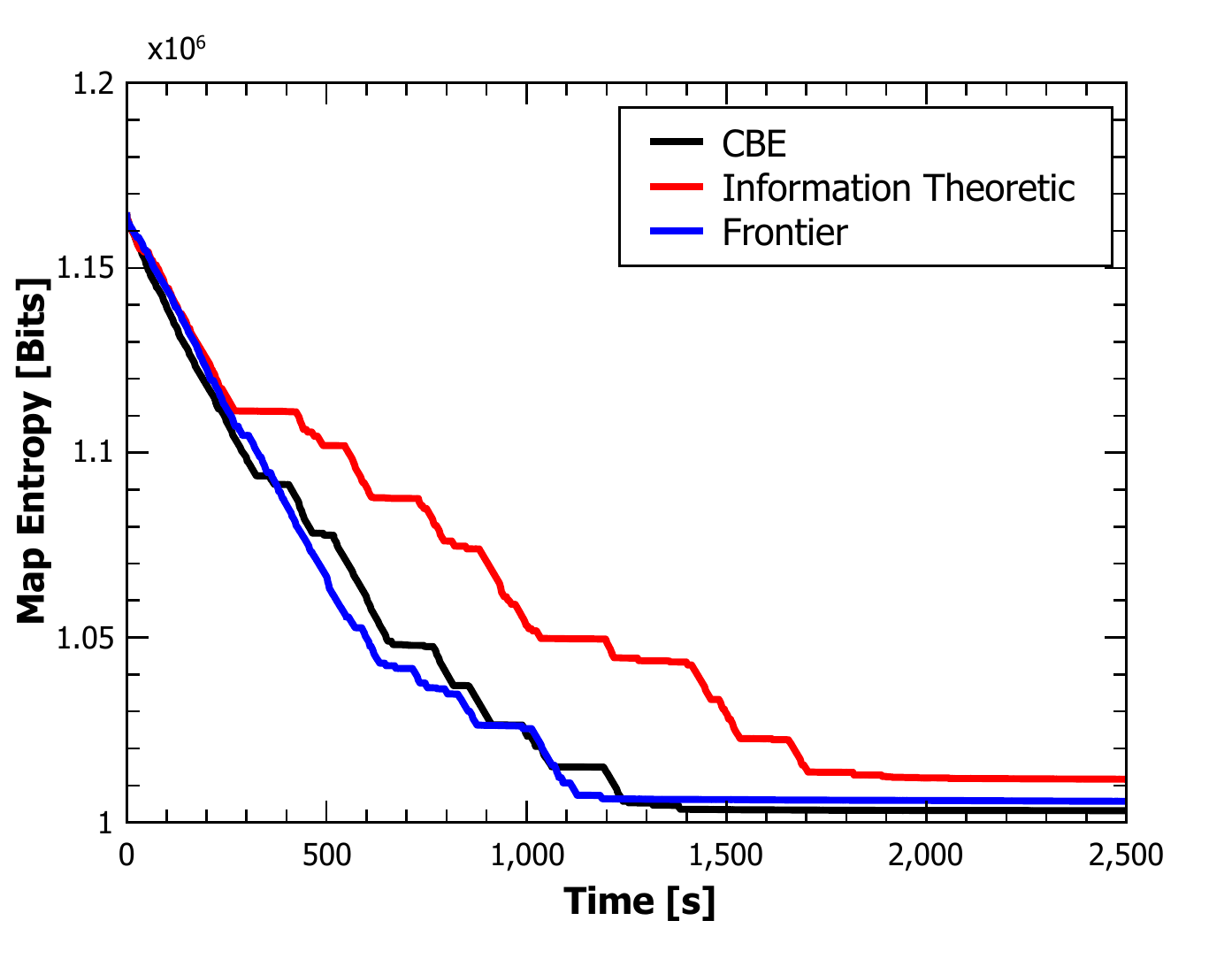}
  	
  	\caption{Jerusalem - Comparison of reduction in map entropy between exploration methods presented in Fig. \ref{JeruslaemMapComparison}. The rate of entropy reduction using CBE and Frontier is similar and outperforms the information-theoretic method, which corresponds to lower instances of crossing already explored regions of the map.}
  	\label{JerusalemComparisonGraph}
  \end{figure}
Fig. \ref{VeniceMapComparison} depicts the executed path of a robot exploring the surroundings of the Piazza San Marco,Venice, while Fig. \ref{JeruslaemMapComparison} shows exploration around the Church of The Holy Sepulchre, Jerusalem. In both cases, the road network is complex, creating many possibilities for autonomous actions. From this qualitative comparisons, one can see that all techniques cover almost the entire mission area (blue square) with no isolated pockets of unexplored regions. However, a closer examination of the executed paths reveals a significant difference. As the map is \textit{a-priori} unknown, it is reasonable that the robot will have to move occasionally through already mapped roads. However, the paths generated by the information-theoretic heuristics revisits known roads much more than the other planners. Although the search heuristics includes a distance penalty, it is not general enough to be effective in all scenarios. Once again, the reason behind such a sub-optimal performance lies in the basic properties of the global point planners i.e. separating the solution for the goal point from the subsequent path generation. We should note that in these scenarios the frontier planner is not as affected. Choosing the closest frontier as the planner's goal point keeps path planning in the robot's local neighbourhood. However, such an arbitrary goal point selection can not always guarantee an optimal path.
   
A quantitative comparison for the Venice and Jerusalem exploration simulations are shown in Figs. \ref{VeniceComparisonGraph} and \ref{JerusalemComparisonGraph}, respectively. Most importantly, even in unfavourable conditions, CBE achieves performance as good as frontier. A careful inspection of the results reveals additional insights. Similarly to the unstructured environment simulations, the initial rate in which the map entropy drops is similar in all techniques. As the robot moves through the map, the number of possible actions increases leading to differences in performance. In Venice, frontier is less effective at first, while in Jerusalem the information-theoretic solution is less effective. Yet, in both experiments the CBE algorithm kept a consistent performance comparable or better than the other leading method regardless which one it is. These results affirm the conclusion from our previous experiment that constrained BO is a robust exploration method. 

The essence of the CBE method lies in its reasoning about the usefulness and safety of the entire path taken. The benefits of reasoning on the overall accumulated reward are more distinct when compared to the information-theoretic single goal point exploration technique (red line). The qualitative results shown in Figs. \ref{VeniceMapComparison} and \ref{JeruslaemMapComparison} expose the inefficiencies in the single goal point methods. The slower exploration rate in this method corresponds to the revisiting of already explored regions of the map whilst moving to the next goal point. By reasoning on the path utility instead of the end goal point, CBE avoids selecting paths that provide little information. Furthermore, the global component in the constrained BO reward function is found to be very effective in pulling robot away from dead-ends. Yet, a more expressive path option is more desirable in such a case, since it will allow longer term planning.

\subsection{Real Environments}

\begin{figure}[htbp]
  	
  	\centering
  	
  	\includegraphics[width=0.45\textwidth]{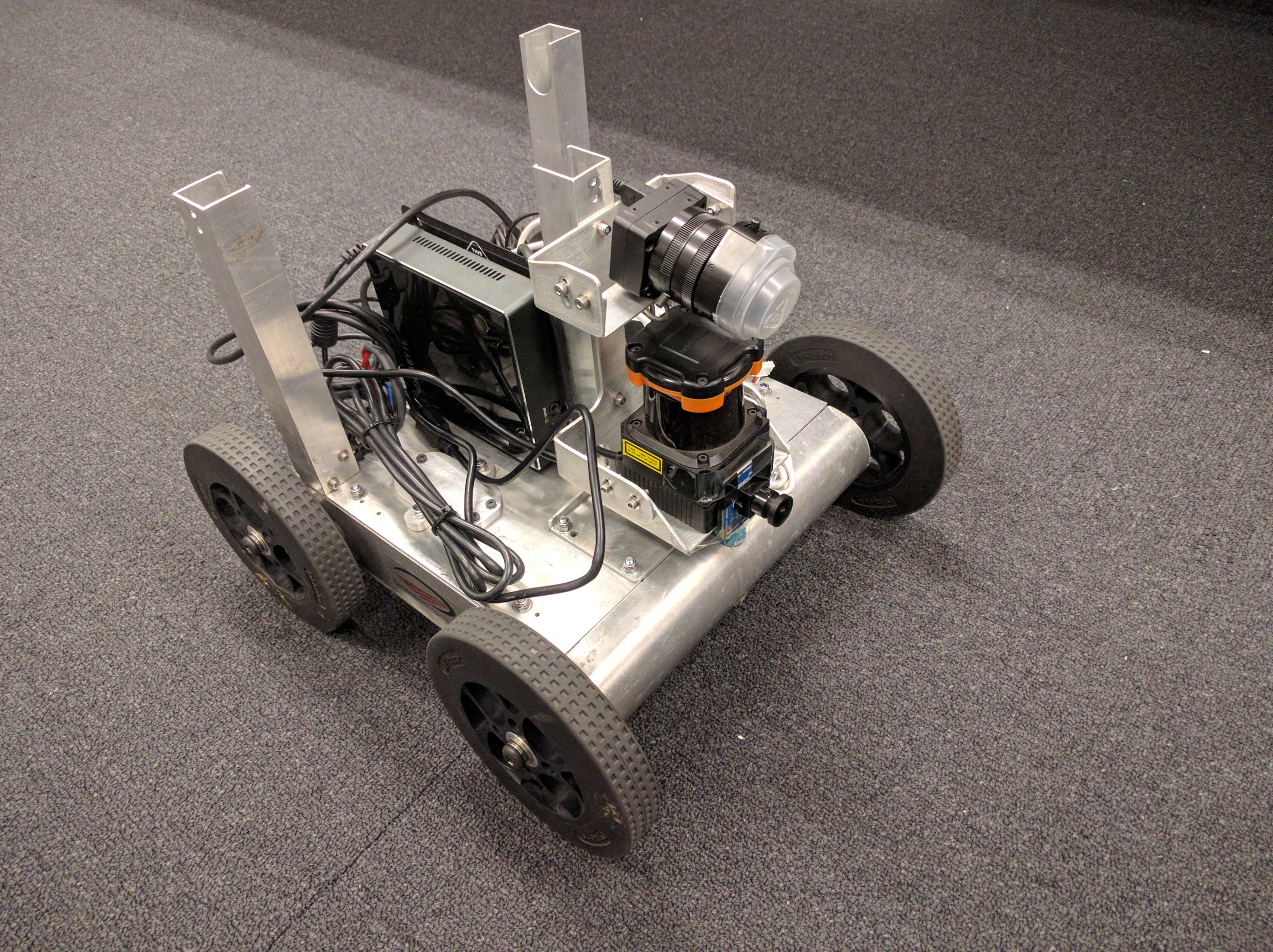}
  	
  	\caption{The Wombot - a mobile robot equipped with a Hokuyo UTM-30lx laser range finder.}
  	\label{fig:Wombot}
  \end{figure}

Simulations show the effectiveness of CBE for autonomous exploration. However, to assess performance with partially observed poses, CBE was evaluated with a real robot mapping a cluttered office environment. We used our in-house robot, the Wombot (see Fig. \ref{fig:Wombot}), equipped with an i7-4500U 1.8GHz dual-core on-board PC and an Hokuyo UTM-30lx laser range finder. 
  
We use \textit{Robot Operating System} (ROS) \citep{quigley2009ros} to manage the communication between the various components of the robot i.e. sensors, actuators etc., and software modules. For mapping and localisation, we utilise externally provided ROS package, gMapping \citep{Grisetti2007}. It is worth noting that the aim of these experiments is to assess the performance in the presence of pose uncertainty. Although CBE can encode loop-closing heuristic in its objective function, we only used the objective function defined in Algorithm \ref{CBE_path_check_uncertain}.

~\begin{figure*}[]
  	
  	\centering
  	
  	\includegraphics[width=0.65\textwidth]{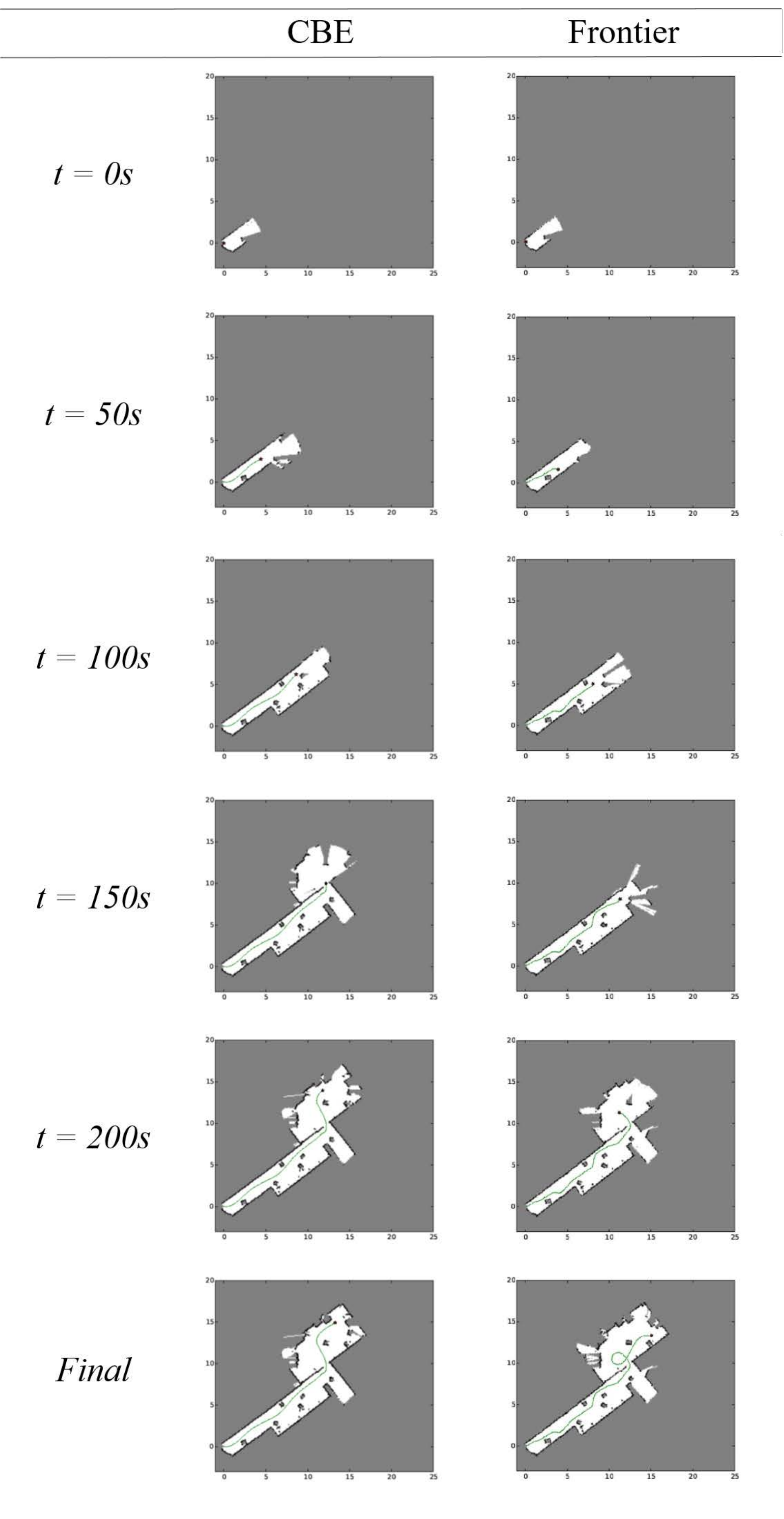}
  	
  	\caption{Map building comparison in a real environment. Each image represents the available map at a specific time stamp shown on the left. The frontier method produces a map less efficiently as it chooses unnecessary long paths due to frontiers forming in occluded space behind the various obstacles.}
  	\label{fig:robot_exp}
  \end{figure*}
Fig. \ref{fig:robot_exp} shows the maps and paths taken at different time stamps for both CBE and frontier. Similarly to experiments in simulations, the frontier planner places a goal point at the closest frontier. Fig. \ref{fig:robot_exp} shows that both methods cover the entire space. However, while both stay clear of obstacles, the path taken by the frontier planner is less efficient. The occluded space behind the various obstacles forms frontiers that are then visited by the robot. As the robot visits these goal points, the overall path length and exploration time increase. CBE, on the other hand, considers the utility of the entire path as opposed to only considering the utility of the goal point. Furthermore, assessing safety using sigma paths that considers the effect of the uncertain pose with robot motion proved to be successful.

\begin{figure}[]
	
	\centering
	
	\includegraphics[width=0.45\textwidth]{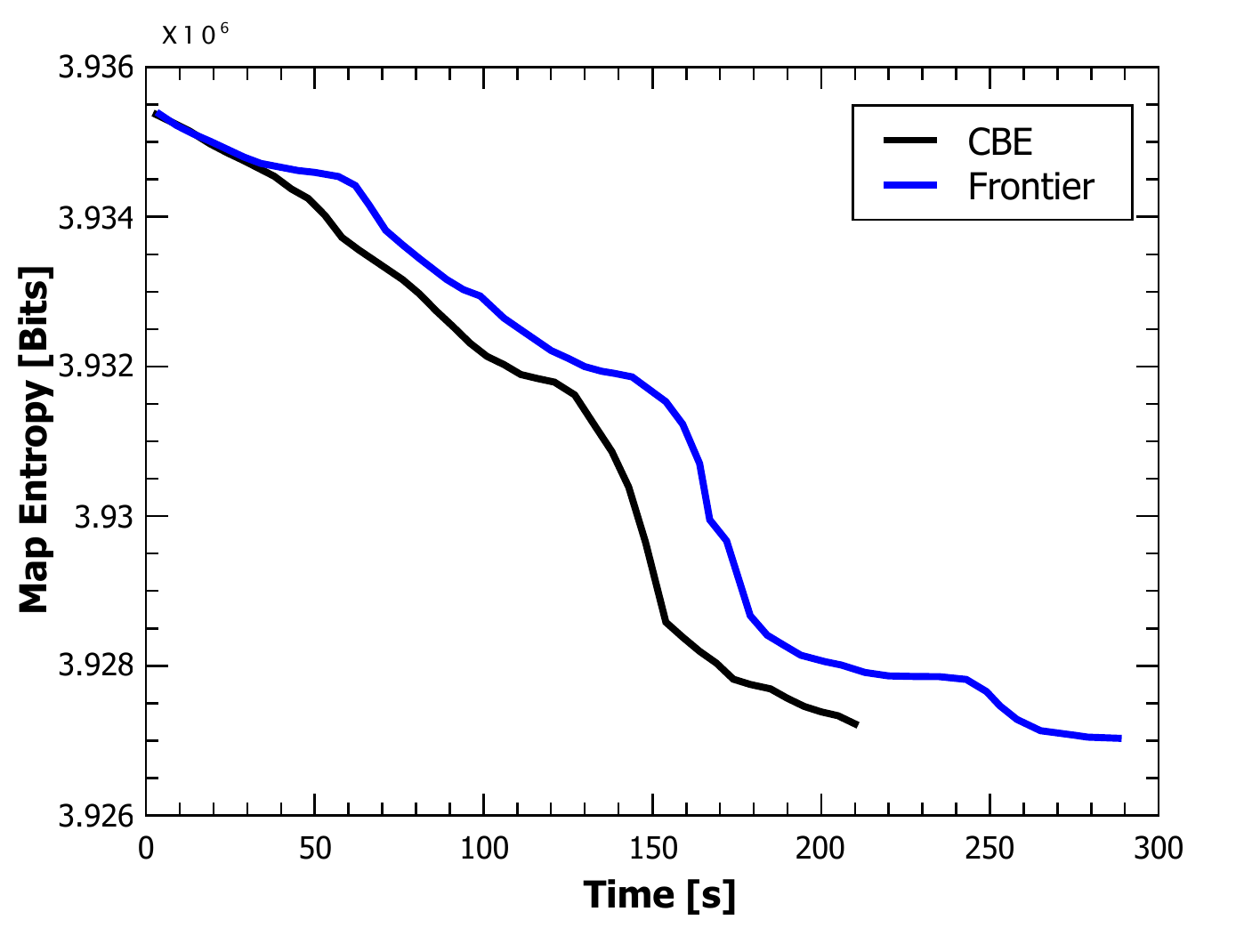}
	\caption{Autonomous exploration with a real robot - Comparison of reduction in map entropy between exploration methods presented in Fig. \ref{fig:robot_exp}; The frontier planner visits the occluded space behind the various obstacles leading to longer path and exploration time. CBE maximises the information gain along its path. Therefore, the path avoids unnecessary maneuvers, which results in an efficient exploration.}
	\label{fig:RobotEtropy}
\end{figure}

Fig. \ref{fig:RobotEtropy} provides a quantitative comparison between the two methods. In the beginning, both methods perform similarly. After about 50 seconds the two methods start to diverge as the frontier planner pulls the robot towards a goal point behind an obstacle. As the exploration continues, the performance gap between the two methods increase, mainly due to a non-optimal goal point selection by the frontier planner. The final map, and the reduction in overall entropy is the same in both methods, although it took frontier roughly 30 seconds longer to do so. 

The advantage of CBE comes with a computational cost as shown in Table \ref{tab:computation_time}. Finding a frontier and planning a safe path from the robot pose to a specific goal point depends of the size of the map and the distance to the goal point. Typically in our experiments, planning took less than 1 second. However, the search for a safe path took longer when there was no safe path to the selected frontier. In such a case, a safe path to a different goal point was calculated, but only after the first search had exhausted its time budget. CBE is more stable despite the higher computational cost. Apart from the first planning instance, which includes a hyperparameters training stage, all the subsequent instances took slightly less than a minute. As shown in section \ref{CBO_exploration}, updating and querying the GPCs is the main the culprit. Choosing a different constraint representation has the potential to dramatically improve CBE's speed.  

\begin{table}[]
	\centering
	\caption{Planning and execution time comparison}
	\label{tab:computation_time}
	\begin{tabular}{|l|c|c|}
		\hline
		& BO   & Frontier \\
		\hline 
		1st plan [s]    & 164  & -                           \\
		Average Plan [s]   & 58.4 & 9.8                          \\
		Execution [s]	& 192 &	236                          \\
		\hline                        
	\end{tabular}
\end{table}

\section{Conclusions}
\label{sec:conclusions}
\subsection{Conclusions}
\label{subsec:conclusion}
This paper proposes a new strategy for safe autonomous exploration. Its novelty lies in the holistic probabilistic approach to robotic exploration. Specifically the paper presents the following contributions:
\begin{inparaenum}[(i)]  
 	
 	\item Formulation of autonomous exploration as a Bayesian optimisation problem in constrained continuous space, where the path is evaluated by its accumulative reward and not only by the reward of its goal point. Traditional exploration methods consist of a two-step solution. First, a collection of goal points (typically one) is defined by a set of heuristics, followed by a path planning step. As a result, the expected usefulness of the resolved path is based solely on the utility function of the end point and does not consider any potential gains along the way. Our new strategy, on the other hand, does not set goal points. Rather, it optimises the path selection by learning the properties of the objective function and any associated constraints. Consequently, the full potential of the robot trajectory can be exploited and not only that of the end point. 
 	
 	\item Constrained Bayesian Exploration as a holistic approach to safe exploration. This method directs the optimisation process in the presence of unknown constraints and risks. Utilising Bayesian inference, the optimiser learns the models of the rewards and constraints. These models are then used to generate a coherent objective function that incorporates gains, costs and risks of any path, allowing efficient identification of potential optimal solutions that satisfy the constraints with high confidence. Unlike traditional exploration techniques, constraints are an integral part of the search mechanism. As the actual model of the constrains is learned online, its incorporation in the optimisation utility is straightforward. Hence, CBE allows a relatively simple and smooth application of other limitations such as energy and time budget. In addition, the probabilistic representation of the constraints can allow the user to balance risk and gain in the process of exploration. 
 \end{inparaenum}
 
To test the efficiency of our method, we compared its performance with other exploration techniques. The results show that the performance of each of the other exploration techniques depends on the layout of the environment. By reasoning on the usefulness of the entire path instead of only its goal point, CBE exhibits a robust and consistent performance in all tests. Even in unfavourable conditions of structured environments, CBE performs better than or as good as the leading method. 

The use of sigma paths to incorporate localisation uncertainty proved successful. A robot travelling through cluttered office space managed to avoid obstacles while still optimising the cost function.

\subsection{Future Work}
\label{subsec:future}
The main disadvantage of CBE is the computational cost. In the algorithm GPCs are the main computational bottleneck as the cost of updating and querying GPCs is cubic in the number of training points. A promising approach to alleviate this restriction is to utilise stochastic variational inference for GPC \citep{hensman2015scalable}, which has a computational cost independent of the the number of data points. 

Another avenue for future work is in extension of the CBE framework to a more flexible family of trajectories. The use of a predefined family of trajectories, such as quadratic or cubic splines, limits the decision space of the robot. This problem is more pronounced when path planning is constrained, for example near obstacles, as the optimisation space is confined. Examples of more expressive path generation already exists in the literature \citep{Yang2013, Charrow-RSS-15}. However, these methods do not consider the overall reward along the path, or only locally optimise the path selection. Combining CBE with an RKHS motion planning method, such as the method presented by \citet{Marinho2016}, may allow for global optimisation of highly expressive paths. 

\bibliography{MyCollectionIJRR}

\end{document}